\documentclass[11pt,twocolumn]{article}

\usepackage[
  a4paper,
  top=0.7in,
  bottom=0.8in,
  left=0.75in,
  right=0.75in
]{geometry}

\usepackage{fontspec}
\usepackage{xeCJK}

\setCJKsansfont[
  BoldFont=FandolHei-Bold.otf
]{FandolHei-Regular.otf}

\usepackage{latexsym}
\usepackage{xcolor}
\usepackage{booktabs}
\usepackage{longtable}
\usepackage{amsfonts}
\usepackage{amsmath}
\usepackage{nicefrac}
\usepackage{microtype}
\usepackage{graphicx}
\usepackage{subcaption}
\usepackage{multirow}
\usepackage{makecell}
\usepackage{diagbox}

\usepackage[authoryear,round]{natbib}
\let\cite\citep
\usepackage[hidelinks]{hyperref}
\usepackage{url}

\usepackage{chessfss}

\graphicspath{{media/}}

\usepackage{titlesec}

\titleformat{\section}
  {\large\bfseries}
  {\thesection}
  {0.6em}
  {}

\titleformat{\subsection}
  {\normalsize\bfseries}
  {\thesubsection}
  {0.6em}
  {}

\usepackage[most]{tcolorbox}
\usepackage{fancyvrb}
\usepackage{fvextra}

\definecolor{gameboxorange}{HTML}{C86A17}
\definecolor{gameboxcream}{HTML}{FFF9F2}
\definecolor{headergray}{HTML}{293138}

\newtcolorbox{gameboxstyle}[1]{
  enhanced,
  breakable,
  colback=gameboxcream,
  colframe=gameboxorange,
  boxrule=0.9pt,
  arc=2.5mm,
  outer arc=2.5mm,
  left=4mm,
  right=4mm,
  top=3mm,
  bottom=3mm,
  colbacktitle=gameboxorange,
  coltitle=white,
  fonttitle=\large\rmfamily,
  title={#1},
  before skip=6pt,
  after skip=6pt
}

\newenvironment{gamebox}[1]
  {\VerbatimEnvironment
   \begin{gameboxstyle}{#1}%
   \begin{Verbatim}[
     fontsize=\small,
     breaklines=true,
     breakanywhere=true
   ]}
  {\end{Verbatim}\end{gameboxstyle}}

\newtcolorbox{pythonboxstyle}[1]{
  enhanced,
  breakable,
  colback=green!3,
  colframe=green!50!black,
  coltitle=white,
  colbacktitle=green!50!black,
  title={#1},
  fonttitle=\bfseries,
  arc=2mm,
  boxrule=0.8pt,
  left=2mm,
  right=2mm,
  top=1mm,
  bottom=1mm
}

\newenvironment{pythonbox}[1]
  {\VerbatimEnvironment
   \begin{pythonboxstyle}{#1}%
   \begin{Verbatim}[
     fontsize=\small,
     breaklines=true,
     breakanywhere=true
   ]}
  {\end{Verbatim}\end{pythonboxstyle}}

\usepackage[colorinlistoftodos]{todonotes}

\newcommand{\cready}[1]{} 

\usepackage{enumitem}
  \setlist{itemsep=0.5pt, topsep=0pt,leftmargin=1.4em,labelsep=0.5em}

\usepackage{hyperref} 
\usepackage{xcolor}
\hypersetup{
    colorlinks,
    linkcolor={red!50!black},
    citecolor={blue!50!black},
    urlcolor={blue!80!black}
}

\usepackage{amsmath}
\usepackage[capitalize,nameinlink]{cleveref}
\AddToHook{cmd/appendix/before}{\crefalias{section}{appendix}}
\Crefname{equation}{Eq.}{Eqs.}
\Crefname{figure}{Fig.}{Figs.}
\Crefname{tabular}{Tab.}{Tabs.}

\usepackage{etoolbox} 
\makeatletter
\pretocmd{\appendix}{%
  \@addtoreset{figure}{section}%
  \@addtoreset{table}{section}%
}{}{}
\makeatother

\usepackage{todonotes}

\begin{document}

\twocolumn[
\begin{@twocolumnfalse}



\noindent
\raisebox{-0.22\height}{%
  \includegraphics[height=1.6em]{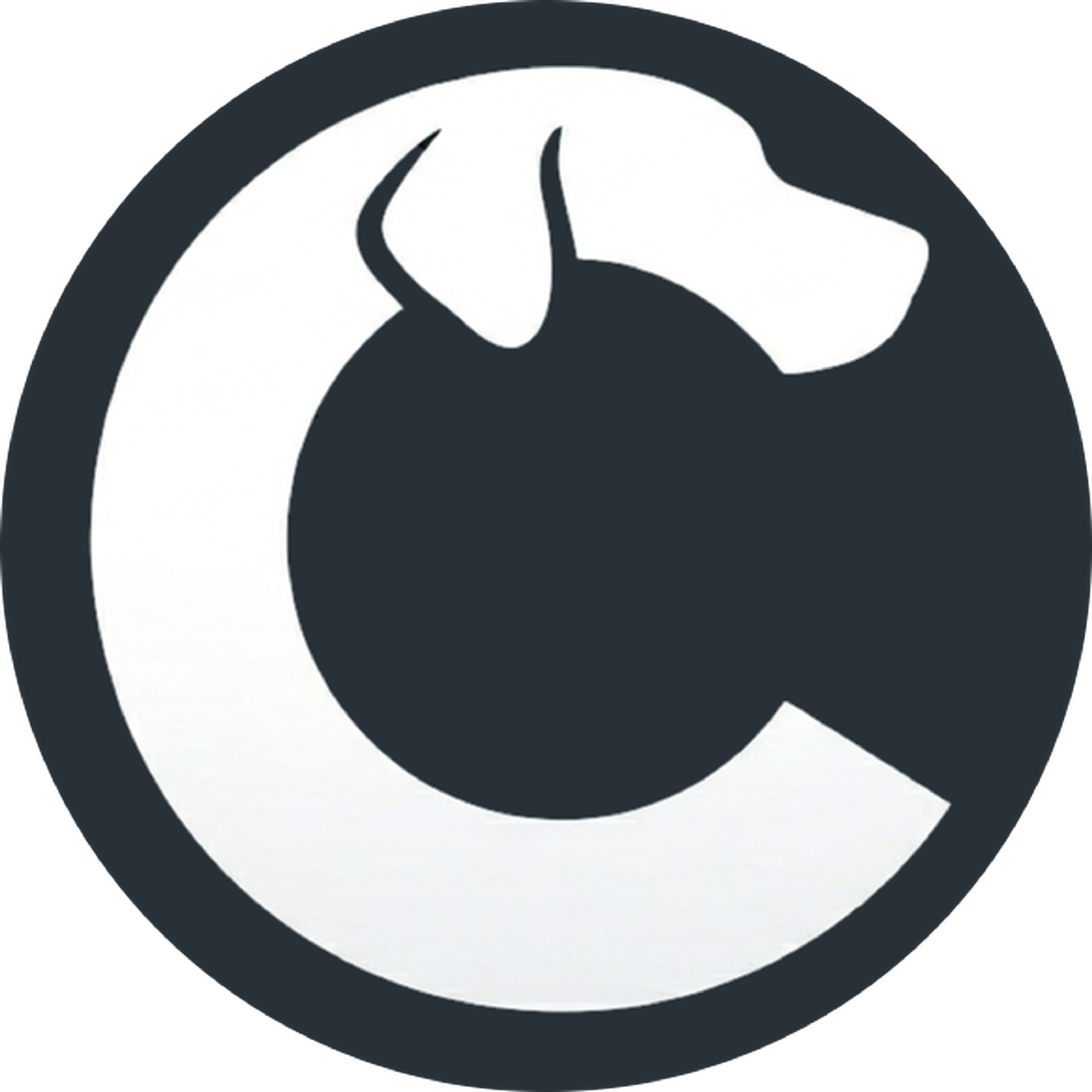}%
}
\hspace{0.4em}%
{\color{headergray}\sffamily\Large\bfseries CoLab Lab}
\hfill
{\color{headergray}\small 2026-08-19}


\noindent
{\color{headergray}\rule{\textwidth}{0.5pt}}

\vspace{1.2em}


{\fontsize{21}{25}\selectfont\bfseries
Skill Issue: Are Skills Language-Invariant in LLMs?
\par}

\vspace{0.9em}







{\fontsize{8.7}{10.2}\selectfont
\mbox{\textbf{Bobby Cheng}$^{\knight,\dagger}$},
\mbox{\textbf{Adam Gaber}$^{\bishop}$},
\mbox{\textbf{Zhengyuan Liu}$^{\knight}$},
\mbox{\textbf{Catherine Arnett}$^{\king}$},
\mbox{\textbf{Omer Goldman}$^{\rook}$},
\mbox{\textbf{Cheston Tan}$^{\knight}$},
\mbox{\textbf{Leshem Choshen}$^{\bishop,\queen,\ast}$}
\par
}

\vspace{0.18em}


{\fontsize{7.8}{9.0}\selectfont
$^{\knight}$A*STAR,
$^{\bishop}$Weizmann Institute of Science,
$^{\queen}$MIT-IBM Watson AI Lab,
$^{\rook}$University of Cambridge,
$^{\king}$EleutherAI
\par
}

\vspace{2.0em}


\noindent
\begin{minipage}{\textwidth}
\small

\textbf{Large language models access knowledge inconsistently across languages, but to what extent do they differ in their skill sets when interacting with different languages? This work quantifies \textbf{\textit{cross-lingual skill inconsistency}} orthogonally from knowledge and general benchmark performance. We do this via multilingual self-play: two instances of the same model compete in a text-based game, each interacting through a different language interface. Since the model, opponent, rules, state space, and available actions remain fixed, this setting isolates the effect of language on the model's realized behavior. We build a multilingual extension to TextArena and evaluate three open-weight models across eight languages and six games covering spatial reasoning, imperfect information, resource allocation, and repeated interaction.\footnotemark We find that the same model can exhibit markedly different playing strength across languages, with systematic variation in win--loss margins, invalid actions, and strategic tendencies. Detailed analyses reveal language-specific failures in spatial reasoning, card-conditioned decisions, and optimal move selection. In some settings, changing only the intermediate reasoning language recovers much of the lost performance, suggesting that language can affect different stages of the decision process. These results show that skill discrepancies are a measurable major roadblock in the development of truly multilingual models. Better understanding these discrepancies can help us design models that perform more equitably across languages.}
\end{minipage}

\vspace{2.0em}

\end{@twocolumnfalse}
]


\begingroup
\renewcommand{\thefootnote}{\fnsymbol{footnote}}

\footnotetext[2]{%
Corresponding author:
\href{mailto:bobbycxy1994@gmail.com}
{\nolinkurl{bobbycxy1994@gmail.com}}%
}

\footnotetext[1]{%
Corresponding author:
\href{mailto:leshem.choshen@mail.huji.ac.il}
{\nolinkurl{leshem.choshen@weizmann.ac.il}}%
}

\endgroup

\footnotetext[1]{%
All the relevant code and data resources are publicly available at
\url{https://github.com/TextArena/TextArena}.%
}


\section{Introduction}

As large language models (LLMs) become more multilingual \cite[e.g.,][]{xue2021mt5massivelymultilingualpretrained, shi2022languagemodelsmultilingualchainofthought} it also becomes clearer that they perform unequally across languages \cite{hu2020xtrememassivelymultilingualmultitask, xcopamultilingualcommonsense, arnett-bergen-2025-language}. Understanding these differences is crucial for deploying multilingual LLMs fairly and reliably, and for identifying what prevents them from functioning as language-agnostic systems.

Much attention has been therefore given to cross-lingual inconsistency, where models respond differently to translations of the same input, but these works mostly focused on the discrepancy in \textit{knowledge} accessibility \cite[e.g.,][]{jiang-etal-2020-x, Qi_2023}, and pointed to the lack of cross-lingual transfer as its source \cite{ifergan2024beneathsurfaceconsistencyexploring, goldman2025eclekticnovelchallengeset}. Equally consequential, and far less understood, is whether models exhibit a different set of \emph{skills} depending on the language through which they interact. In other words, beyond retrieving different knowledge, can the same model reason, plan, and make decisions more effectively in some languages than in others? Answering this question requires isolating the effect of language on the model's behavior independently of its stored knowledge or overall benchmark performance.

\begin{figure*}[!t]
  \centering
  \includegraphics[width=\textwidth]{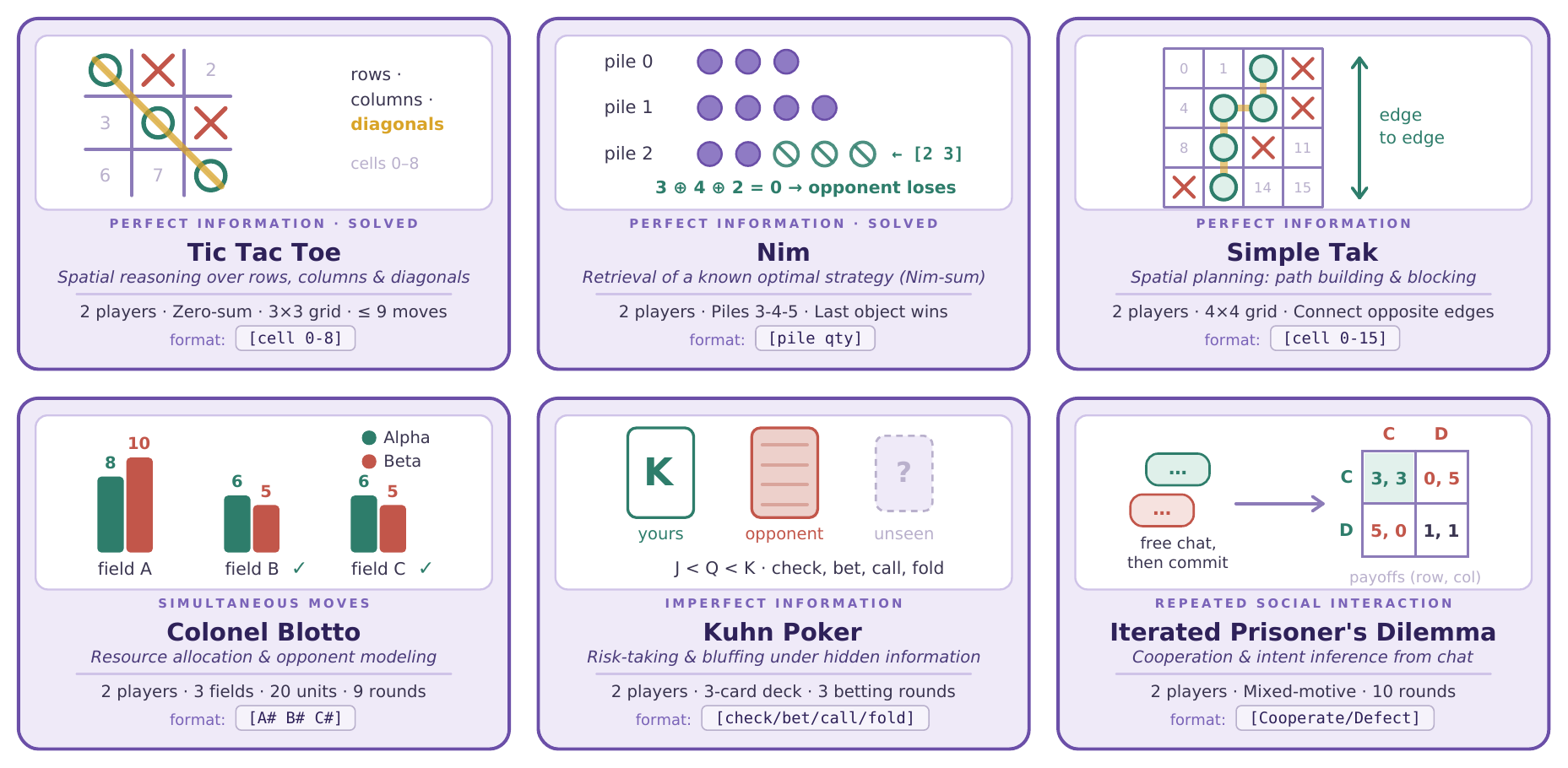}
  \caption{The six game environments and the primary skill each probes; full rules in App.~\ref{app:games}.
  }
  \label{fig:game-example}
\end{figure*}

\begin{figure*}[!t]
    \centering

    \begin{subfigure}[t]{0.95\textwidth}
        \centering
        \includegraphics[width=\linewidth]{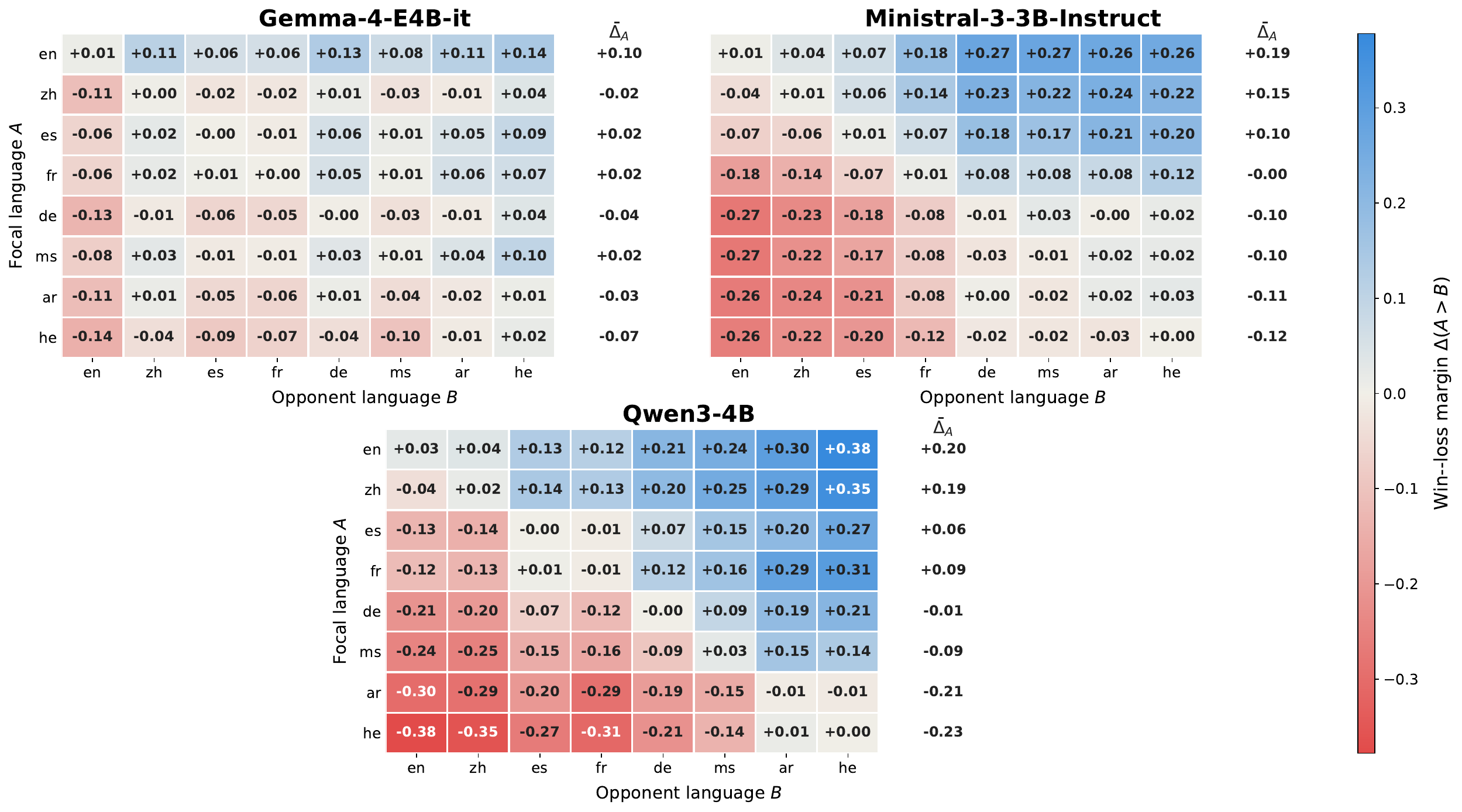}
    \end{subfigure}

    \vspace{0.6em}

    \caption{
    Overall, wins of each language against others, per model. Each heatmap shows the role-pooled win--loss margin $\Delta(A>B)=(W_A-L_A)/N$ aggregated over all games, where positive values mean row language $A$ outperforms column language $B$. The rightmost column reports each language's mean margin $\bar{\Delta}_A$. Diagonal is same language games, with randomly assigned sides for comparison. Across models, English is consistently strong and Hebrew weak, with Qwen3-4B showing the sharpest hierarchy.
    }
    \label{fig:model_level_pairwise_language_margins}
\end{figure*}

\begin{figure}[t]
  \centering
    \includegraphics[width=\columnwidth]{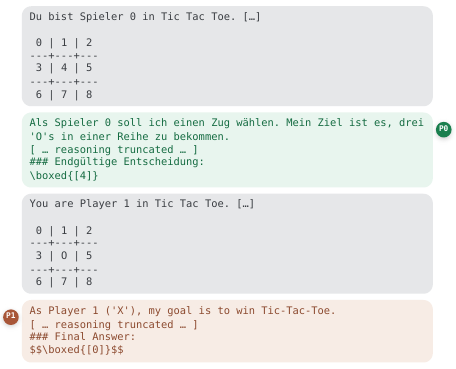}
  \caption{
    Illustration of Gemma-4-E4B-it playing TicTacToe with itself in German and English respectively. 
  }
  \label{fig:chat-example}
\end{figure}

In this work we study {\it cross-lingual skill inconsistency} by letting models compete against themselves in multilingual text-based games. We let two instances of the same model interact via translations of the same environment into different languages, while the board states, cards, numerical information, action spaces, and game rules remain fixed. Fig.~\ref{fig:chat-example} demonstrates a typical game played in German and English. If the model accesses and expresses the same underlying skills through both languages, then the two model instances should exhibit equal playing strengths and the wins and losses should distribute randomly as is the case when playing against oneself in the same language (see Fig.~\ref{fig:model_level_pairwise_language_margins}'s diagonals). 

To support this and future studies, we introduce a large-scale multilingual extension to TextArena (\citealp{guertler2025textarena}; see Sec.~\ref{sec:multilingualtextarena}), \textbf{comprising 65 single-, two-, and multi-player games in 193 languages} of different tiers of translation and providing broad coverage for studying language effects in agentic gameplay. 
Out of these, we experiment with a subset of manually verified translations of six games into eight languages
(see Fig.~\ref{fig:game-example}). 
We use this subset to explore cross-lingual skill inconsistency of three 4B-sized open-weights models: Gemma 4, Ministral 3, and Qwen~3. 

We find LLMs to be behaviorally inconsistent across languages (see Sec.~\ref{sec:different_capabilities}, and Fig. \ref{fig:model_level_pairwise_language_margins}). These differences appear in spatial reasoning, where the axis of failure varies by language; in strategic behavior, where different languages induce different risk profiles and error rates; and in unequal access to known knowledge, where models may retrieve or execute a known strategy reliably in one language but not another (see Sec.~\ref{sec:skill_differences}). We also observed that reasoning in a stronger language can recover substantial performance in some games, suggesting that language sensitivity may arise during reasoning, state interpretation, or both (see Sec.~\ref{sec:recovery}). Finally, we show that these differences are partly explainable by multilingual benchmarks and agree with the uneven availability of languages across training data (see Sec.~\ref{sec:benchmarks_data}).





\section{Multilingual TextArena}
\label{sec:multilingualtextarena}


\begin{table}[t]
\centering
\footnotesize
\setlength{\tabcolsep}{2.5pt}
\renewcommand{\arraystretch}{0.92}
\begin{tabular*}{\columnwidth}{
    @{\extracolsep{\fill}}
    l cccccccc
    @{}
}
\toprule
\textbf{Tier}
& \multicolumn{7}{c}{\textbf{Resource class}}
& \textbf{Total} \\
\cmidrule(lr){2-8}
& 5 & 4 & 3 & 2 & 1 & 0 & -- & \\
\midrule
A & 6 & 0  & 0  & 2  & 0  & 0  & 0 & 8   \\
B & 1 & 17 & 18 & 2  & 4  & 0  & 0 & 42  \\
C & 7 & 1  & 6  & 14 & 76 & 15 & 5 & 124 \\
E & 0 & 0  & 0  & 0  & 10 & 5  & 3 & 18  \\
\midrule
\textbf{Total}
& 14 & 18 & 24 & 18 & 90 & 20 & 8 & 192 \\
\bottomrule
\end{tabular*}
\caption{
Languages by verification tier and resource class.
We focus on the eight Tier-A languages.
}
\label{tab:langs_stats}
\end{table}

\paragraph{TextArena} TextArena~\cite{guertler2025textarena} is an open-source collection of 100+ competitive single-, two-, and multi-player text-based games for training and evaluating LLMs, released under the MIT License. We extend relevant games with multilingual support which allow players' observations to be rendered in a different language without altering the rules, legal actions, rewards or transition dynamics. Examples of the game templates and starter code are in App.~\ref{app:games}.

\paragraph{Translation workflow.} 
To translate TextArena games we used a multi-tiered translation and verification pipeline. Prioritizing diversity in typology, script, culture, and number of speakers, we chose eight languages as our Tier A: English, Arabic, German, Spanish, French, Hebrew, Malay, and Chinese. This selection ensures that at least two languages share a characteristic along each dimension. We translated six TextArena games to these languages with Claude Opus 4.8~\cite{anthropic2026claudeopus48} or GPT-5.2~\cite{singh2026openaigpt5card} (see App.~\ref{app:translationflow} for details), and assigned native speakers to manually verify the translations. This is the set of games and languages that were used in our experiments.

For an additional 42 high and mid resource languages, Tier B, we constructed an automatic translation and verification pipeline. Due to budget constraints, we used the open-weights models of Llama-3.1-405B \cite{grattafiori2024llama3herdmodels} and Qwen2.5-72B \cite{qwen25} to translate 65 games into these languages. Following \citet{dobler2026multilingualreasoninggymmultilingual}, we back-translated the results to English again and let Claude Opus 4.8 to judge whether the back translations were faithful to the original English game. If the procedure failed to produce a faithful translation across four different seeds, the language was demoted to Tier C.

For the remaining 142 languages, where high-quality machine translations are harder to come by, we used NLLB-200 \cite{costa2022no} for translation and back-translation. Llama-3.1-405B and Qwen2.5-72B then independently assessed fidelity. If they both agree that the back-translation was faithful to the original text in 85\% of the time, then the language is assigned to Tier C; otherwise Tier E.


Tab.~\ref{tab:langs_stats} details the number of languages in each tier and in each resourcefulness class (from \citealp{joshi2020state}). See App. \ref{app:localization} for further details.

\section{Experiment Setup}
\label{sec:exp-setup}


\paragraph{Evaluated Models.}

We focus on Gemma-4-E4B-it~\cite{gemmateam2026gemma4technicalreport}, Qwen3-4B~\cite{qwen3technicalreport}, and Ministral3-3B-Instruct-2512~\cite{liu2026ministral3}, which are comparably sized open-weight models that remain tractable for large-scale multilingual self-play.
To determine the general multilingual capability of these models, we evaluate them on Global-MMLU~\cite{singh2024globalmmluunderstandingaddressing} and Belebele~\cite{bandarkar-etal-2024-belebele}, and compare against our multilingual setup in Sec.~\ref{sec:benchmarks_data}.



\paragraph{Game Environments.}
Models were evaluated on six two-player games from TextArena spanning perfect-information (TicTacToe, Nim, and SimpleTak), simultaneous resource allocation (Colonel Blotto), imperfect-information (Kuhn Poker), and repeated social interaction game (Iterated Prisoner's Dilemma). Together, they test spatial and numerical reasoning, planning, allocation, bluffing, cooperation, and adaptation. More details of each game are covered in App.~\ref{app:games}.

\paragraph{Two-player multi-turn games.}
Each game $g$ is implemented as an environment $E_g$ that unfolds over multiple turns according to a Markov transition process. At step $t$, $E_g$ is in state $s_t$, and the active player receives an observation $o_t$ containing the player-visible portion of the interaction history, including all publicly observable past actions $a_{<t}$. The player then selects an action $a_t$, after which $E_g$ transitions to $s_{t+1}$. A trajectory is denoted $\tau=(s_0,o_0,a_0,s_1,o_1,a_1,\ldots,s_H)$, where $s_H$ is a terminal state corresponding to a win, loss, or draw. The outcomes for Player~0 and Player~1 are denoted $r_0(\tau)$ and $r_1(\tau)$, respectively. In standard competitive games, a win for one player is a loss for the other, while a draw yields zero outcome for both. In repeated social-interaction games such as Iterated Prisoner's Dilemma, final outcomes are instead determined by $E_g$'s cumulative scoring rule. If a model submits an invalid action, it is given an opportunity to correct it; otherwise, the game terminates as an immediate loss.

\paragraph{Language assignment.}
For each game $g$, Player~0 receives the game instructions and observations in language $\ell_0$ and Player~1 in language $\ell_1$, where $\ell_i\in\mathcal{L}=\{$English (en), Arabic (ar), German (de), Spanish (es), French (fr), Hebrew (he), Malay (ms), Chinese (zh)$\}$. A language assignment is the ordered pair $(\ell_0,\ell_1)$. Where communication is relevant, each player observes the opponent's messages in their original language, so a model may process both its interface language and the opponent's communication language. A small number of language-independent strings are intentionally preserved across translations to maintain shared game mechanics and action formats. These include game-board symbols and coordinates, card ranks and suits, numerical values, and fixed action syntax such as ``[bet]''; for example, TicTacToe displays available moves as bracketed indices such as ``[1]''.

\paragraph{Action format and sampling parameters.}
We instruct model $m$ via a system prompt to place its final action inside \texttt{\textbackslash boxed\{\}} so that the environment can reliably extract the submitted action (see App.~\ref{app:prompt_templates}). During generation, tokens are sampled autoregressively using \texttt{temperature} $=1.0$, \texttt{top\_p} $=0.95$, and \texttt{top\_k} $=64$ to encourage diverse self-play trajectories. 

\paragraph{Evaluation protocol.}
For each model $m$ and game $g$, we evaluate every language pair, including same-language pairs, in both player-role assignments, with $n=400$ self-play games per direction. With eight languages, this gives
$\left(\binom{8}{2}+8\right)\times 2\times 400=28{,}800$
games per model--game pair, or \textbf{86{,}400 games per game} across three models and \textbf{518{,}400 games overall} across six games. Evaluating both role assignments balances languages across player roles and avoids conflating language effects with structural advantages such as Player~0's first-move advantage in TicTacToe.

\section{Metrics}

\paragraph{Role-pooled win--loss margin.}
For each model $m$, game $g$, and pair of languages $A$ and $B$, we
evaluate both role assignments, $(A,B)$ and $(B,A)$. We pool outcomes
from the perspective of language $A$: a Player~0 win in $(A,B)$ and a
Player~0 loss in $(B,A)$ both count as wins for $A$, while the converse
outcomes count as losses for $A$. Let $W_{m,g}(A,B)$ and
$L_{m,g}(A,B)$ denote these pooled win and loss counts, and let
$N_{m,g}(A,B)$ denote the total number of games across both
assignments. We define the role-pooled win--loss margin as

\begin{equation}
\Delta_{m,g}(A,B)
=
\frac{W_{m,g}(A,B)-L_{m,g}(A,B)}
     {N_{m,g}(A,B)}.
\end{equation}

The margin lies in $[-1,1]$. Positive values indicate that language
$A$ outperforms language $B$, negative values indicate the reverse,
and draws contribute zero. Pooling both assignments controls for
structural player-role advantages. By construction,
$\Delta_{m,g}(A,B)=-\Delta_{m,g}(B,A)$.

\paragraph{Mean language margin.}
We summarize the strength of language $A$ for model $m$ in game $g$ by
averaging its margin against every other language:

\begin{equation}
\mu_{m,g}(A)
=
\frac{1}{|\mathcal{L}|-1}
\sum_{B\in\mathcal{L}\setminus\{A\}}
\Delta_{m,g}(A,B).
\end{equation}

Higher values indicate stronger average self-play performance through
language $A$.

\paragraph{Model-level mean language margin.}
We summarize the overall strength of language $A$ for model $m$ by
macro-averaging its mean margin across games:

\begin{equation}
\bar{\mu}_{m}(A)
=
\frac{1}{|\mathcal{G}|}
\sum_{g\in\mathcal{G}}
\mu_{m,g}(A).
\end{equation}

\section{Results}
\label{sec:results}


In the following sections, we present our main findings. Unless otherwise specified, Gemma, Qwen, and Ministral refer to their 4B-sized models.

\subsection{Capabilities Differ Across Languages}
\label{sec:different_capabilities}
Fig.~\ref{fig:model_level_pairwise_language_margins} shows that the same model can express substantially different capabilities depending on the language through which the game is presented, which we refer to as the \emph{language interface}. This is striking because the games are largely abstract: the boards, cards, numerical information, legal actions, and underlying strategies remain unchanged. Yet English is strongest on average across all three models, while Hebrew is consistently among the weakest. The magnitude of this effect is also model-dependent: Gemma is comparatively stable, whereas Qwen exhibits the sharpest language hierarchy. These results show that even when the core task information is non-linguistic, the language interface can affect which capabilities a model successfully expresses.

\begin{table}[h]
\centering
\footnotesize
\setlength{\tabcolsep}{2.5pt}
\renewcommand{\arraystretch}{1.02}
\begin{tabular*}{\columnwidth}{
    @{\extracolsep{\fill}}
    lccccccc
    @{}
}
\toprule
\textbf{Model}
& \textbf{Blotto}
& \textbf{Nim}
& \textbf{TTT}
& \textbf{Tak}
& \textbf{IPD}
& \textbf{Kuhn}
& \textbf{Avg.} \\
\midrule
Gemma
& 0.44
& 0.02
& \textbf{0.47}
& 0.43
& 0.25
& 0.05
& 0.28 \\
Qwen
& \textbf{1.48}
& \textbf{0.85}
& 0.32
& \textbf{0.44}
& 0.03
& 0.13
& 0.54 \\
Ministral
& 1.28
& 0.49
& 0.36
& 0.28
& \textbf{0.74}
& \textbf{0.20}
& \textbf{0.56} \\
\midrule
Avg.
& 1.07
& 0.45
& 0.38
& 0.38
& 0.34
& 0.13
& 0.46 \\
\bottomrule
\end{tabular*}
\caption{
Language sensitivity by game, measured as language gap, $\max_{\ell}\mu_{\ell}-\min_{\ell}\mu_{\ell}$.
\textbf{Bold} marks most language-sensitive model for each game.
Bottom row shows the average across models.
}
\label{tab:game-language-sensitivity}
\end{table}




Language sensitivity also varies across games (see Tab.~\ref{tab:game-language-sensitivity}). Colonel Blotto shows the largest language gap across all three models, whereas Kuhn Poker is consistently among the least sensitive. The remaining games show more model-dependent effects. For example, Iterated Prisoner's Dilemma is relatively stable for Gemma and Qwen but is more sensitive for Ministral. We next summarise how these differences manifest across specific skills like spatial reasoning, strategic behavior and access to known strategies.

\subsection{Language-Conditioned Skills Difference}
\label{sec:skill_differences}
Beyond aggregate game-playing strength, the language interface produces consistent differences in how models fail. In spatial games like TicTacToe and SimpleTak, the failure mode varies by language. We see that English interfaces tend to show a relatively balanced distribution of losses across rows, columns and diagonals, whereas non-Latin-script interfaces such as Arabic and Hebrew skew consistently toward column and diagonal defeats, with game trajectories revealing frequent mislabeling of cell sequences as lines. 
Strategic behavior shifts as well. In Kuhn Poker, the same model adopts different risk profiles depending on the interface language. Bluffing rates with the weakest card vary by more than twofold across languages for Qwen, while Gemma shows its largest cross-language variation when deciding how to play the intermediate card Q. More broadly, both the magnitude and the type of these language-conditioned shifts differ across models.

The starkest effect concerns access to pre-existing knowledge. Nim admits a complete algorithmic solution (reducing the Nim-sum to zero), which lets us test whether a known strategy is retrievable through each language interface. Mentions of the optimal strategy, optimal first-move execution, and win rates all drop sharply in Arabic and Hebrew for Qwen and Ministral, and a majority of the remaining strategy mentions in these languages originate from the small fraction of games in which the model spontaneously switched into a Latin script mid-reasoning. This indicates that the strategy is present in the model but not reliably accessible through every language: merely changing the processing language can retrieve knowledge that would otherwise be lost. Full per-game analyses, including defeat distributions, card-conditioned action rates, and the Nim strategy results, are provided in App.~\ref{app:skill_analyses}.

\subsection{Stronger Languages Enable Recovery}
\label{sec:recovery} 

\begin{table}[h]
\centering
\footnotesize
\renewcommand{\arraystretch}{1.15}

\begin{tabular*}{\linewidth}{
    @{\extracolsep{\fill}}
    l
    c
    c
    c
    @{}
}
\toprule
\textbf{Game}
& \textbf{Floor}
& \textbf{Reasoning language}
& \textbf{Ceiling} \\
& $\mu_{\text{weak}}$
& $\mu$ / \textbf{recovery}
& $\mu_{\text{best}}$ \\
\midrule

Kuhn
& $-0.03$ {\scriptsize zh/zh}
& \begin{tabular}[c]{@{}c@{}}
    $-0.01$ {\scriptsize zh/es} / $37.6\%$ \\
    $-0.01$ {\scriptsize zh/fr} / $49.3\%$
  \end{tabular}
& $+0.02$ {\scriptsize es/es} \\

\addlinespace[2pt]

ST
& $-0.21$ {\scriptsize de/de}
& \begin{tabular}[c]{@{}c@{}}
    $+0.05$ {\scriptsize de/en} / $60.5\%$ \\
    $-0.16$ {\scriptsize de/es} / $11.6\%$
  \end{tabular}
& $+0.22$ {\scriptsize en/en} \\

\addlinespace[2pt]

TTT
& $-0.22$ {\scriptsize de/de}
& \begin{tabular}[c]{@{}c@{}}
    $+0.20$ {\scriptsize de/en} / $89.4\%$ \\
    $-0.14$ {\scriptsize de/es} / $17.0\%$
  \end{tabular}
& $+0.25$ {\scriptsize en/en} \\

\bottomrule
\end{tabular*}

\caption{
Role-corrected strength $\mu$ under each interface/reasoning language pair.
The middle column reports the strongest reasoning language, followed by the
second strongest, while holding the weak interface language fixed.
Recovery is
$(\mu-\mu_{\text{weak}})/
(\mu_{\text{best}}-\mu_{\text{weak}})$.
Kuhn, ST and TTT denote Kuhn Poker, SimpleTak and TicTacToe.
}
\label{tab:interface_reasoning_language}
\end{table}

Prior works show that multilingual models can sometimes improve performance on non-English inputs by routing them through English. For example, self-translation and question-alignment methods improve multilingual task performance by translating non-English inputs into English before inference or reasoning~\cite{zhu-etal-2024-question, etxaniz-etal-2024-multilingual,mondshine2025beyond}. Here, we separate the language of the environment interface from the language of the intermediate reasoning and found that reasoning in a stronger language can recover some of the performance.

Tab.~\ref{tab:interface_reasoning_language} shows that German is Gemma's weakest interface language in TicTacToe, with $\mu=-0.22$. Holding the German interface fixed while switching the reasoning language to English increases the margin to $\mu=+0.20$, recovering $89.4\%$ of the reachable gap to the English-interface ceiling. We observe a similar effect in SimpleTak, where switching from German to English reasoning improves the margin from $\mu=-0.21$ to $\mu=+0.05$, recovering $60.5\%$ of the reachable gap. Because these gains occur without translating or replacing the environment observations, they suggest that a substantial portion of the language effect in these games arises during the model's intermediate reasoning process. However, recovery is limited and non-monotonic in Kuhn Poker, indicating that language sensitivity can arise at different stages of the agent's decision process and cannot always be addressed by reasoning in a stronger language.

\begin{figure*}[!t]
\centering

\begin{subfigure}[t]{0.95\textwidth}
    \centering
    \includegraphics[width=\linewidth]
        {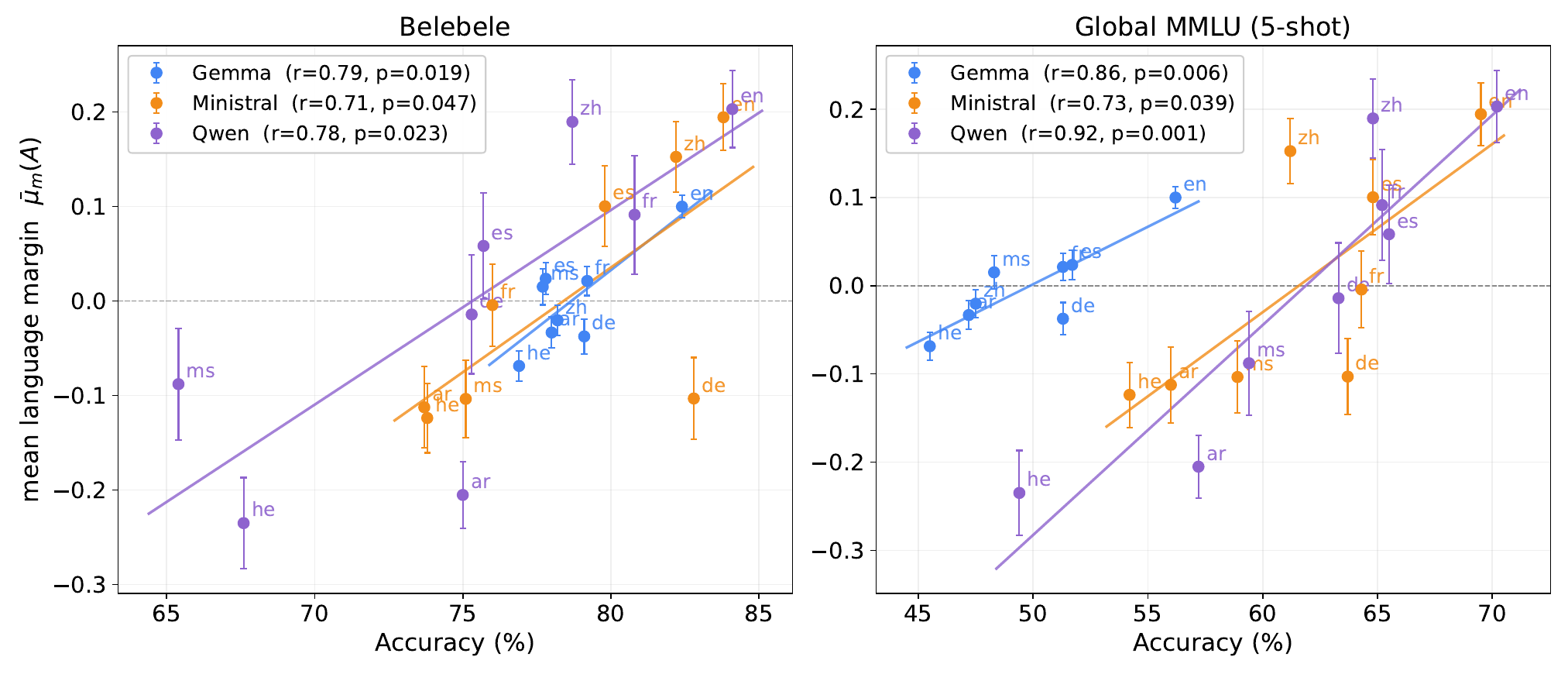}
    \caption{
    Mean language margin against Belebele accuracy (left) and
    Global MMLU accuracy (right).
    }
    \label{fig:margin_vs_benchmarks}
\end{subfigure}

\vspace{0.5em}

\begin{subfigure}[t]{0.95\textwidth}
    \centering
    \includegraphics[width=\linewidth]
        {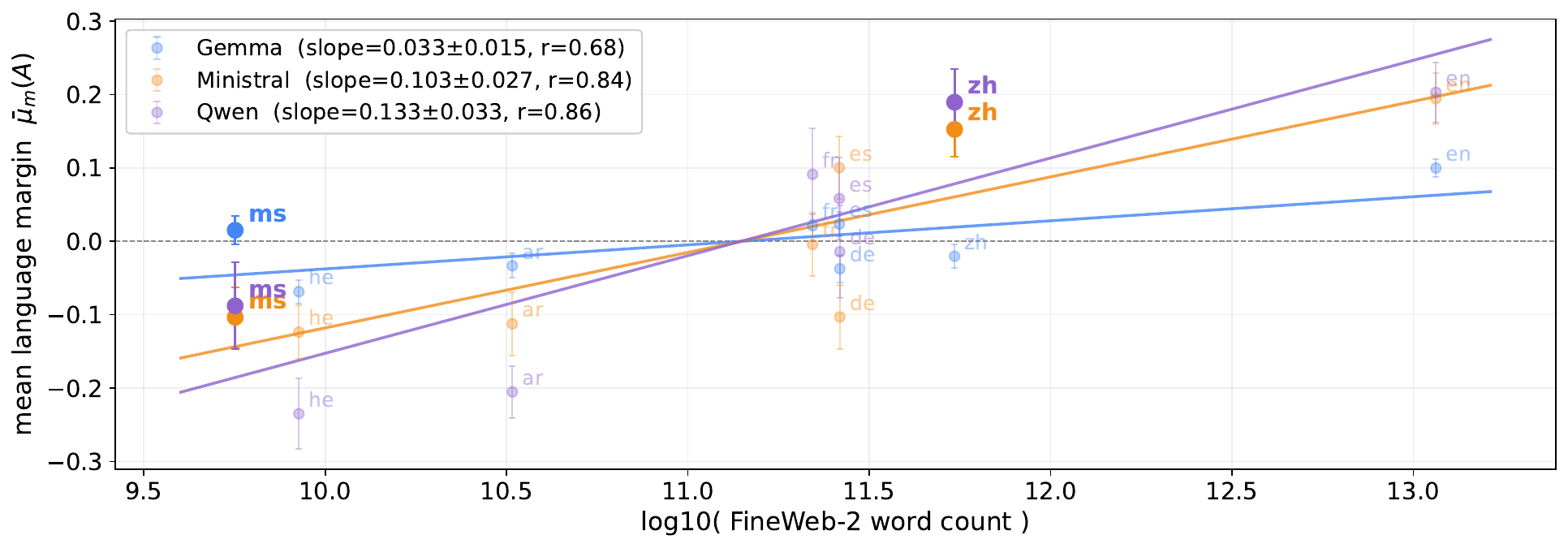}
    \caption{
    Mean language margin against web-text availability, measured by
    $\log_{10}$ FineWeb-2 word count, with English estimated from FineWeb.
    }
    \label{fig:margin_vs_data}
\end{subfigure}

\caption{
Relationship between within-model language strength and external measures of
language capability and data availability. Each point represents one of eight
languages for a given model, lines show per-model least-squares fits, and
error bars show standard errors across the language's seven pairwise margins.
In \subref{fig:margin_vs_benchmarks}, Pearson $r$ is reported in the legend
($n=8$). In \subref{fig:margin_vs_data}, highlighted points mark the largest
deviations from the fitted trends. Margins are comparable across languages
within, but not across, models.
}
\label{fig:margin_external_correlates}
\end{figure*}

\subsection{Explaining Outcome Differences}
\label{sec:benchmarks_data}


Finally, we ask whether the language-conditioned outcome differences can be
explained by two external references: each model's static multilingual
competence, and the amount of multilingual text available on the web.

\begin{table}[h]
\centering
\footnotesize
\setlength{\tabcolsep}{3.2pt}
\renewcommand{\arraystretch}{0.94}
\begin{tabular}{@{}lccc@{}}
\toprule
\textbf{Lang.}
& \makecell{\textbf{Gemma-4}\\\textbf{E4B-it}}
& \textbf{Qwen3-4B}
& \textbf{Ministral3-3B} \\
\midrule

\multicolumn{4}{@{}c}{\textit{Belebele}} \\
Arabic  & \textbf{78.0} & 75.0 & 73.7 \\
German  & 79.1 & 75.3 & \textbf{82.8} \\
English & 82.4 & \textbf{84.1} & 83.8 \\
Spanish & 77.8 & 75.7 & \textbf{79.8} \\
French  & 79.2 & \textbf{80.8} & 76.0 \\
Hebrew  & \textbf{76.9} & 67.6 & 73.8 \\
Malay   & \textbf{77.7} & 65.4 & 75.1 \\
Chinese & 78.2 & 78.7 & \textbf{82.2} \\
\textit{Avg.}
& $\mathbf{78.7{\scriptstyle\pm1.7}}$
& $75.3{\scriptstyle\pm6.3}$
& $\mathbf{78.4{\scriptstyle\pm4.2}}$ \\

\midrule

\multicolumn{4}{@{}c}{\textit{Global MMLU}} \\
Arabic  & 47.2 & \textbf{57.2} & 56.0 \\
German  & 51.3 & 63.3 & \textbf{63.7} \\
English & 56.2 & \textbf{70.2} & 69.5 \\
Spanish & 51.7 & \textbf{65.5} & 64.8 \\
French  & 51.3 & \textbf{65.2} & 64.3 \\
Hebrew  & 45.5 & 49.4 & \textbf{54.2} \\
Malay   & 48.3 & \textbf{59.4} & 58.9 \\
Chinese & 47.5 & \textbf{64.8} & 61.2 \\
\textit{Avg.}
& $49.9{\scriptstyle\pm3.4}$
& $\mathbf{61.9{\scriptstyle\pm6.4}}$
& $\mathbf{61.6{\scriptstyle\pm5.0}}$ \\

\bottomrule
\end{tabular}
\caption{
Accuracy (\%) on Belebele and 5-shot Global MMLU.
Bold indicates the best-performing models for each language and benchmark
average.
}
\label{tab:multilingual-knowledge-results}
\end{table}

\paragraph{Static Benchmarks} In spite there being only a sample size of 8 languages, there is noticeable correlation between the mean language margin and Global MMLU (5-shot) at $r$ between $0.73$ and $0.92$ depending on the model, and likewise with Belebele at $r$ between $0.71$ and $0.79$ (see Fig.~\ref{fig:margin_vs_benchmarks}). The language ranking observed in interactive play is thus partly explainable from static benchmarks. Languages a model scores well on in isolation are also, on average, the languages it wins with in self-play. What benchmark competence does \emph{not} predict, however, is stability across language which is the max--min range of a model's mean language margins. Gemma has the lowest Global MMLU mean of the three models yet the smallest cross-language spread, whereas Qwen has the highest mean and the largest spread. Benchmark level and cross-lingual consistency are therefore distinct axes, and being better on average does not imply behaving more uniformly across languages.

\paragraph{Data Availability}

Since the training data distributions of the models we evaluate are not publicly available, we cannot directly measure how much text each model was exposed to in each language. We therefore use the amount of publicly available web text as a proxy for language-level data availability, using per-language word counts from FineWeb-2~\cite{penedo2025fineweb2pipelinescale}. For English, which is not included in FineWeb-2, we estimate the corresponding word count from FineWeb~\cite{penedo2024the} using an English token-to-word fertility of approximately $1.3$, consistent with values reported in FineWeb-2. We then relate each model's mean language margin $\bar{\mu}_m(A)$ to this proxy (see Fig.~\ref{fig:margin_vs_data}). The fit within each model is positive and strong, with $r$ averaging $0.79$. Languages with more available web text tend to be stronger interfaces, mirroring the benchmark correlations above.



The more informative pattern lies in the deviations from this trend, highlighted in Fig.~\ref{fig:margin_vs_data}. Malay sits above the pooled fit for all three models despite having the smallest FineWeb-2 corpus, and in particular outperforms Hebrew for every model even though Hebrew has more available web text by this proxy. This asymmetry is consistent with ECLeKTic's finding that cross-lingual transfer is stronger between languages sharing a writing system, and with prior work identifying script as a key factor in cross-lingual knowledge and skill transfer~\cite{goldman2025eclekticnovelchallengeset, ifergan2024beneathsurfaceconsistencyexploring, malkin-etal-2022-balanced, mittal2023mokb6multilingualopenknowledge, diskind2026cross}. This suggests that Latin-script Malay may benefit from transfer from higher-resource Latin-script languages, while Hebrew cannot exploit the same script-based transfer.

Chinese exposes the opposite limitation of a pure data account. Despite having roughly $20\times$ less available web text than English by this proxy, Qwen and Ministral reach near-English margins, while Gemma's Chinese margin remains near zero. The same data availability thus supports very different realized strength depending on the model. Moreover, since Chinese achieves this without sharing a script with the other strong interfaces, script alone cannot account for the results either. Data quantity and script each explain part of the language hierarchy, but the remaining variation is model-specific: how much strength a model realizes from a given language is not determined by how prevalent that language is in web data.

\section{Related Works}\label{sec:related_work}



\subsection{Multilingual Language and Reasoning Evaluation}

Multilingual benchmarks have progressed from evaluating language understanding tasks such as natural language inference, question answering, and commonsense reasoning~\cite{ponti-etal-2020-xcopa, lin-etal-2022-shot}, to covering broader capabilities including mathematical reasoning, reading comprehension, knowledge-intensive academic tasks, code generation and instruction following~\cite{Shi2022LanguageMA, bandarkar-etal-2024-belebele, singh2024globalmmluunderstandingaddressing, huang-etal-2025-benchmax}, with recent benchmarks evaluating regional and culturally situated knowledge by drawing questions from local sources ~\cite{romanou2024includeevaluatingmultilinguallanguage, mrl-workshop-2025-global-piqa}.

These benchmarks primarily evaluate models through fixed inputs and predefined answers. They reveal whether accuracy varies across languages, but not whether a model expresses an equivalent policy over an evolving interaction. We use Belebele and Global-MMLU as reference for general multilingual competence (see Tab.~\ref{tab:multilingual-knowledge-results}), while Multilingual TextArena studies whether a fixed model accesses and expresses the same interactive skills when operating in the same environment through different language interfaces.

\subsection{Cross-Lingual Knowledge and Skill Transfer}


Prior work suggests that multilingual models exhibit factual compartmentalization, where information acquired through one language might be irretrievable in another ~\cite{goldman2025eclekticnovelchallengeset, asai2021xorqacrosslingualopenretrieval, chua2025crosslingualcapabilitiesknowledgebarriers, limkonchotiwat-etal-2022-cl, litschko2025crossdialectinformationretrievalinformation}. Some offer post-hoc solutions, often following inference in multiple languages \citep{huang-etal-2023-languages, diskind2026cross}. Even knowledge that seems to be consistent across languages is often shown to be stored twice rather than shared \cite{ifergan2024beneathsurfaceconsistencyexploring,Qi_2023}. At the same time, learning linguistic skills is cheap, with 100M parameter models showing equivalent performance to state-of-the-art 70B ones~\cite{charpentier2025findings,chang2026goldfish}. 
Together, these findings suggest that factual compartmentalization is more than a discrepancy in linguistic ability between languages within the same model\cready{cite Adam's}.

Related work on skill transfer studies whether task competence acquired from supervision in one or few languages generalizes to others~\cite{hu2020xtrememassivelymultilingualmultitask, malkin-etal-2022-balanced, shaham2024multilingualinstructiontuningjust}. For example, \citet{turc2021revisitingprimacyenglishzeroshot} shows that the sourced language used for fine-tuning affects zero-shot transfer performance, with English not always providing the strongest transfer to other languages.

Our work is related to cross-lingual skill transfer, but differs from the standard source-fine-tuning setup.  
Instead, we focus on language-conditioned skill access and expression rather than acquiring a skill through one language.

\subsection{Interactive and Game-Based Evaluation of LLMs}

Interactive benchmarks evaluate LLMs as agents whose actions affect an evolving environment. Unlike static benchmarks, these settings test whether models can interpret changing states, select valid actions, and adapt over multiple turns. These include tool-use  environments and game-based benchmarks that evaluate capabilities such as planning, spatial reasoning, learning from interaction, coordination, and goal-directed decision-making
~\cite{barres2025tau2,guertler2025textarena,
wu2024smartplaybenchmarkllmsintelligent,
gong2023mindagentemergentgaminginteraction,
qiao2023gameevalevaluatingllmsconversational}.

A related line of work uses strategic and game-theoretic environments to evaluate planning, decision-making, and social reasoning, including direct competition between LLMs
~\cite{duan2024gtbenchuncoveringstrategicreasoning,
costarelli2024gamebenchevaluatingstrategicreasoning,
yao2025spinbenchllmsplanstrategically}.

Within this line of work, existing benchmarks primarily compare models, prompting methods, or agent architectures under a shared or fixed language interface. We instead use competitive environments to study language-conditioned variation \emph{within} a fixed model by extending TextArena with multilingual interfaces. 

\section{Conclusion}

In this paper, we introduced Multilingual TextArena and used controlled self-play to test whether the skills of an LLM remain consistent across language interfaces. Across three models, eight languages and six games, we found differences in their playing strength, strategic behavior and spatial features within the same model.

Reasoning in a stronger language recovered substantial performance in some games, but these were limited in other games, suggesting that language can affect multiple stages of interaction, including state interpretation, reasoning, knowledge retrieval, and action selection. Static multilingual benchmarks and relative web-data availability explained part of the variation, not all.

Overall, we showed that a \textit{skill} a model has may not be equally applied in every language. Multilingual evaluation should assess not only whether models understand equivalent inputs, but also if they \textit{behave} consistently across languages.
\section*{Limitations}

\paragraph{Closed-data models} 

While the models we evaluated have their technical reports published, their training data remains behind closed doors. This closed-data nature makes it hard for us to uncover what reasons and training approaches might have explained the difference in language performance per model. While there were models like Apertus~\cite{swissai2025apertus} which state clearly their pre-training data recipe and their multilingual distribution, Apertus produced several invalid moves which made it difficult to produce a viable score. Sticking with these models, the explainability of these results required assumptions and estimates, e.g. availability of multilingual data. 

\paragraph{Model scale.}

Our evaluation is limited to models in the 3B to 4B parameter range. This scale enabled controlled and cost-efficient evaluation over more than half a million games. However, cross-lingual skill inconsistency may differ for larger models. Our results, then, should therefore not be interpreted as establishing scale-invariant language effects.




\section*{Acknowledgements}
We would like to thank the Shimon and Golde Picker--Weizmann Annual Grant and the Center for New Scientists at the Weizmann Institute of Science for supporting this research. Omer Goldman also acknowledges support from the Blavatnik Family Foundation.


\bibliographystyle{acl_natbib}
\bibliography{references}


\appendix
\section{Prompt Templates}
\label{app:prompt_templates}

We use model-specific chat wrappers to match each model's expected input
format, but keep the task instruction as consistent as possible across models.
The scientifically relevant prompt variants are the default action prompt and the
language-conditioned reasoning prompt used in our intervention experiments.


\paragraph{Default action prompt.}
This prompt is used when the model is asked to play the game without an
explicit instruction to reason in the provided language.
\begin{quote}
\small
You are a competitive game player. Make sure you read the game instructions
carefully, and put your final answer within \texttt{\textbackslash boxed\{\}}.
\end{quote}

\paragraph{Language-conditioned reasoning prompt.}
This prompt is used in the main multilingual experiments. It instructs the
model to reason in the language provided by the environment interface.
\begin{quote}
\small
You are a competitive game player. Make sure you read the game instructions
carefully, reason in the language provided, and put your final answer within
\texttt{\textbackslash boxed\{\}}.
\end{quote}

\section{Experimental Setup}
\label{app:experimentalsetup}
Our experiments involve inference only; no model training or hyperparameter search was performed. All models are served with vLLM~\cite{kwon2023efficient} and orchestrated with Ray~\cite{moritz2018raydistributedframeworkemerging} for distributed rollout collection.

A primary experimental run evaluates one model \(m\) on one game \(g\) across
all eight languages and comprises 28,800 self-play games. Each run uses two
NVIDIA H200 GPUs. Across three models and six games, the primary evaluation
comprises 18 runs and 518,400 games. Excluding Iterated Prisoner's Dilemma,
each primary run required approximately 6 H200 GPU-hours.

For Global-MMLU, we used the implementation provided in the LM-Evaluation-Harness~\cite{eval-harness} repository and retained its default task
configuration and evaluation parameters. For Belebele~\cite{bandarkar-etal-2024-belebele}, we used the official benchmark repository.

\section{Translation Workflow}
\label{app:translationflow}
Concretely, we used a one-shot prompt to extend each game with multilingual support. Claude~Opus~4.8~\cite{anthropic2026claudeopus48} was given an original monolingual environment, a multilingual extension of a reference environment, and the corresponding English template. It was then prompted to generate the multilingual implementation and English template for a new game, given the target environment in its original monolingual form.

After validating this process across several games, we translated the resulting English templates using GPT-5.2~\cite{singh2026openaigpt5card}. Google Translate was used to produce back-translations, and native speakers reviewed their fidelity to the intended game interactions. Manual review by three native speakers rarely identified substantive issues.

\section{Game Environments}
\label{app:games}

TextArena~\cite{guertler2025textarena} follows the OpenAI Gym (now Gymnasium)~\cite{brockman2016openaigym} interface, enabling easy use and extension. For multilingual games, each player's language is specified through a language mapping passed to the environment's \texttt{reset} method, as illustrated below.

\begin{pythonbox}{Multilingual game playing illustration}
import textarena as ta 

# initialize the players
agents = {
    0: ta.agents.OpenRouterAgent("google/gemma-4-31b-it"),
    1: ta.agents.OpenRouterAgent("google/gemma-4-31b-it"),
}

# initialize the environment
env = ta.make(env_id="TicTacToe-v0") 
env.reset(num_players=len(agents), lang_mapping={0: "he", 1: "en"})

# main game loop
done = False 
while not done:
    player_id, observation = env.get_observation()
    action = agents[player_id](observation)
    done, step_info = env.step(action=action)
rewards, game_info = env.close()
print(rewards)
print(game_info)
\end{pythonbox}

\subsection{Nim}
\label{app:nim}

Nim~\cite{bouton1901} is a two-player strategy game played with several piles of objects. Players take turns removing one or more objects from a single pile. Under the standard rules, the player who removes the final object wins. The game requires players to reason about the configuration of the remaining piles and select moves that leave the opponent in a disadvantageous position.

\paragraph{Action format.} Moves are submitted as \texttt{[pile quantity]}, where \texttt{pile} indexes one of the piles and \texttt{quantity} is the number of objects to remove; e.g., \texttt{[0 3]} removes three objects from pile 0.

Examples of the starting game prompt:

\begin{gamebox}{Hebrew Translation}
[משחק] ברוך הבא לנים, שחקן 0!
חוקים:
- בתורך, הסר לפחות חפץ אחד מערימה אחת בלבד.
- השתמש בפורמט '[ערימה כמות]' כדי להסיר חפצים, לדוגמה '[0 3]'.
- מי שלוקח את החפץ האחרון מנצח!
[משחק] ערימת האבנים הנוכחית:
  ערימה 0: 3
  ערימה 1: 4
  ערימה 2: 5
\end{gamebox}

\begin{gamebox}{Spanish Translation}
[JUEGO] ¡Bienvenido a Nim, Jugador 0!
Reglas:
- En tu turno, elimina al menos un objeto de exactamente una pila.
- Usa el formato '[pila cantidad]' para eliminar objetos, por ejemplo '[0 3]'.
- ¡Quien tome el/los último(s) objeto(s) gana!
[JUEGO] Pila actual:
  pila 0: 3
  pila 1: 4
  pila 2: 5
\end{gamebox}

\subsection{Tic Tac Toe}
\label{app:ttt}

Tic Tac Toe~\cite{weisstein_tictactoe} is a two-player game played on a ($3 \times 3$) grid. Players alternate placing their symbol, either ($X$) or ($O$), in an empty cell. The first player to form a horizontal, vertical, or diagonal line of three symbols wins. If the grid is filled without either player completing such a line, the game ends in a draw.

\paragraph{Action format.} Moves are submitted as \texttt{[cell]}, where \texttt{cell} is the index (0--8) of an empty square; e.g., \texttt{[4]} places the player's mark in the center cell.

Examples of the starting game prompt:

\begin{gamebox}{Chinese Translation}
[游戏] 你是井字棋游戏中的玩家 1。
你的目标是在棋盘上连成三个（横向、纵向或对角线）。
轮到你时，请选择一个格子编号（0-8）来放置你的标记。
例如，'[4]' 将你的标记放在棋盘中央格子。
作为玩家 1，你的标记是 'X'，对手的标记是 'O'。
[游戏] 当前棋盘：

 0 | 1 | 2 
---+---+---
 3 | 4 | 5 
---+---+---
 6 | 7 | 8 

可用落子位置：'[0]', '[1]', '[2]', '[3]', '[4]', '[5]', '[6]', '[7]', '[8]'
\end{gamebox}

\begin{gamebox}{Hebrew Translation}
[משחק] אתה שחקן 0 במשחק איקס עיגול.
המטרה שלך היא להשיג שלושה ברצף (אופקית, אנכית או באלכסון) על הלוח.
בתורך, בחר את מספר התא (0-8) שבו תרצה להציב את הסימן שלך.
לדוגמה, '[4]' מציב את הסימן שלך בתא המרכזי.
בתור שחקן 0, הסימן שלך הוא 'O', והיריב שלך הוא 'X'.
[משחק] הלוח הנוכחי:

 0 | 1 | 2 
---+---+---
 3 | 4 | 5 
---+---+---
 6 | 7 | 8 

מהלכים זמינים: '[0]', '[1]', '[2]', '[3]', '[4]', '[5]', '[6]', '[7]', '[8]'
\end{gamebox}

\subsection{Colonel Blotto}
\label{app:cb}

Colonel Blotto~\cite{colonel_blotto} is a two-player resource-allocation game played across several battlefields. Each player simultaneously distributes a fixed number of troops among the battlefields. A battlefield is won by the player who assigns more troops to it, while equal allocations result in a tie. The player who wins the most battlefields wins the game.

\paragraph{Action format.} Allocations are submitted as \texttt{[A\# B\# C\#]}, assigning a non-negative number of units to each field; the amounts must sum to exactly 20. E.g., \texttt{[A7 B7 C6]}.

Examples of the starting game prompt:

\begin{gamebox}{Malay Translation}
[PERMAINAN] Anda ialah Komander Alpha dalam permainan Colonel Blotto. Dalam setiap pusingan, anda mesti mengagihkan tepat 20 unit merentasi medan berikut: A, B, C
Format: '[A7 B7 C6]'
Menangi majoriti medan untuk memenangi pusingan!
[PERMAINAN] === COLONEL BLOTTO - Pusingan 1/9 ===
Pusingan dimenangi - Komander Alpha: 0, Komander Beta: 0
\end{gamebox}

\begin{gamebox}{English Translation}
[GAME] You are Commander Alpha in a game of ColonelBlotto. Each round, you have to allocate exactly 20 units across fields: A, B, C
Format: '[A7 B7 C6]'
Win the majority of fields to win the round!
[GAME] === COLONEL BLOTTO - Round 1/9 ===
Rounds Won - Commander Alpha: 0, Commander Beta: 0    
\end{gamebox}

\subsection{Kuhn Poker}
\label{app:kp}
Kuhn Poker~\cite{Kuhn1951} is a simplified two-player poker game using only three cards: a Jack, Queen, and King. Each player is dealt one card, while the remaining card is hidden. Players then complete a single betting round in which they may check, bet, call, or fold. If neither player folds, the player holding the higher card wins.

\paragraph{Action format.} Moves are submitted as one of \texttt{[check]}, \texttt{[bet]}, \texttt{[call]}, or \texttt{[fold]}, restricted to the actions legal at the current point of the betting sequence.

Examples of the starting game prompt:

\begin{gamebox}{French Translation}
[JEU] Vous êtes le joueur 1 dans une partie de Kuhn Poker en 3 manches.
Règles du jeu :
- Le Kuhn Poker utilise un jeu de 3 cartes : J, Q, K (J est la plus faible, K la plus forte)
- Chaque joueur mise 1 jeton en entrée et reçoit 1 carte à chaque manche (remarque : les cartes sont distribuées sans remise, vous ne pouvez donc pas avoir la même carte que votre adversaire)
- La partie se déroule sur 3 manches
- Le joueur ayant le plus de jetons à la fin de toutes les manches gagne
Règles des actions :
- '[check]' : Passer sans miser (uniquement si aucune mise n’est en cours)
- '[bet]' : Ajouter 1 jeton au pot (uniquement si aucune mise n’est en cours)
- '[call]' : Suivre la mise de l’adversaire en ajoutant 1 jeton
- '[fold]' : Se coucher et laisser l’adversaire remporter le pot
[JEU] ### Début de la manche 1 sur 3. Votre carte est : 'J'
[JEU] Vos actions disponibles sont : '[check]', '[bet]'
\end{gamebox}

\begin{gamebox}{German Translation}
[SPIEL] Du bist Spieler 1 in einem 3-Runden-Spiel von Kuhn Poker.
Spielregeln:
- Kuhn Poker verwendet ein 3-Karten-Deck mit J, Q, K (J ist die niedrigste, K die höchste Karte)
- Jeder Spieler zahlt 1 Chip als Einsatz und erhält in jeder Runde 1 Karte (Hinweis: Die Karten werden ohne Zurücklegen ausgeteilt, daher kannst du nicht dieselbe Karte wie dein Gegner haben)
- Das Spiel geht über 3 Runden
- Der Spieler mit den meisten Chips nach allen Runden gewinnt
Aktionsregeln:
- '[check]': Passen ohne zu setzen (nur wenn kein Einsatz auf dem Tisch liegt)
- '[bet]': 1 Chip zum Pot hinzufügen (nur wenn kein Einsatz auf dem Tisch liegt)
- '[call]': Einen gegnerischen Einsatz mit 1 Chip ausgleichen
- '[fold]': Deine Hand aufgeben und dem Gegner den Pot überlassen
[SPIEL] ### Runde 1 von 3 beginnt. Deine Karte ist: 'K'
[SPIEL] Deine verfügbaren Aktionen sind: '[check]', '[bet]'
\end{gamebox}

\subsection{Simple Tak}
\label{app:st}

Simple Tak~\cite{Rothfuss2011} is a simplified two-player connection game played on a square grid. Players take turns placing pieces on empty spaces, with the objective of forming a continuous path connecting two opposite sides of the board. Unlike standard Tak, pieces cannot be stacked or moved after placement. A player may therefore either extend their own path or block spaces needed by their opponent.

\paragraph{Action format.} Moves are submitted as \texttt{[cell]}, where \texttt{cell} is the index (0--15) of an empty cell; e.g., \texttt{[12]} places the player's stone in cell 12.

Examples of the starting game prompt:

\begin{gamebox}{Arabic Translation}
[اللعبة] أنت اللاعب 0 في SimpleTak.
على اللوحة، تظهر أحجارك بالرمز 'O' وتظهر أحجار خصمك بالرمز 'X'.

في دورك، اختر خانة فارغة واحدة باستخدام رقمها، وضع حجرك فيها.
على سبيل المثال، '[12]' يضع حجرك في الخانة 12.

هدفك هو تكوين مسار متصل من أحجارك يربط بين حافتين متقابلتين من اللوحة، إما من الأعلى إلى الأسفل أو من اليسار إلى اليمين.
[اللعبة] حالة اللوحة الحالية:

+----+----+----+----+
| 0  | 1  | 2  | 3  |
+----+----+----+----+
| 4  | 5  | 6  | 7  |
+----+----+----+----+
| 8  | 9  | 10 | 11 |
+----+----+----+----+
| 12 | 13 | 14 | 15 |
+----+----+----+----+
الحركات المتاحة: [0], [1], [2], [3], [4], [5], [6], [7], [8], [9], [10], [11], [12], [13], [14], [15]
\end{gamebox}

\begin{gamebox}{English Translation}
[GAME] You are Player 0 in SimpleTak.
On the board, your stones appear as 'O' and your opponent's stones appear as 'X'.

On your turn, choose one empty cell (by its numbered index) and place your stone there.
For example, '[12]' places your stone in cell 12.

Your objective is to form a continuous path of your stones that connects two opposite edges of the board (top-to-bottom or left-to-right).
[GAME] Current Board:

+----+----+----+----+
| 0  | 1  | 2  | 3  |
+----+----+----+----+
| 4  | 5  | 6  | 7  |
+----+----+----+----+
| 8  | 9  | 10 | 11 |
+----+----+----+----+
| 12 | 13 | 14 | 15 |
+----+----+----+----+
Available Moves: [0], [1], [2], [3], [4], [5], [6], [7], [8], [9], [10], [11], [12], [13], [14], [15]
\end{gamebox}

\subsection{Iterated Prisoners Dilemma}
\label{app:ipd}
The Iterated Prisoner's Dilemma~\cite{Axelrod84} is a repeated two-player mixed-motive game. In each round, both players simultaneously choose to cooperate or defect: mutual cooperation yields 3 points each, mutual defection 1 point each, and unilateral defection yields 5 points to the defector and 0 to the cooperator. Before each decision, players exchange free-form messages over a fixed number of communication turns, allowing negotiation, promises, and deception. The game spans 10 rounds, and the player with the higher cumulative score wins.

\paragraph{Action format.} During communication turns, players exchange free-form text. In the decision phase, the message must contain \texttt{[Cooperate]} or \texttt{[Defect]}.

Examples of the starting game prompt:

\begin{gamebox}{English Translation}
[GAME] You are Player 0 in an Iterated Prisoner's Dilemma spanning 10 rounds.

Game Structure:
- Before each decision you have 1 turns to communicate freely.
- After that, both players simultaneously choose to cooperate or defect.

Payoff Matrix (fixed each round):
- Both Cooperate -> each 3
- Both Defect -> each 1
- One Defects, one Cooperates ➜ Defector 5, Cooperator 0

How to Play:
- During conversation: type any text you wish.
- During decision phase: include '[Cooperate]' or '[Defect]' (case-insensitive). You may add extra text before/after the token.
[GAME] --- Starting Round 1 ---
\end{gamebox}

\begin{gamebox}{Chinese Translation}
[游戏] 你是玩家 0，正在进行一场持续 10 轮的重复囚徒困境游戏。

游戏结构：
- 在每次决策之前，你有 1 个回合可以自由交流。
- 之后，双方玩家同时选择合作或背叛。

收益矩阵，每轮固定：
- 双方合作 -> 每人获得 3
- 双方背叛 -> 每人获得 1
- 一方背叛，一方合作 ➜ 背叛者获得 5，合作者获得 0

玩法说明：
- 在交流阶段：输入任何你想说的文字。
- 在决策阶段：包含 '[Cooperate]' 或 '[Defect]'，不区分大小写。你可以在该标记前后添加额外文本。
[游戏] --- 第 1 轮开始 ---
\end{gamebox}


\section{Multilingual UI Localization}
\label{app:localization}

The languages used in this paper's experiments are localized with the manual,
native-speaker-reviewed workflow of Sec.~\ref{sec:multilingualtextarena}. This
appendix documents the \emph{separate}, fully automatic pipeline, with only
sporadic manual verification, behind the released multilingual resource. It
retains that workflow's structural guarantees but drops systematic native review
in order to scale across the resource spectrum. We contribute upstream
adaptations to the open-source TextArena project~\citep{guertler2025textarena},
making $65$ games and one shared UI file translatable ($64$ locale files, roughly
$1{,}000$ player-facing strings carrying about $1{,}100$
\texttt{\{placeholder\}} slots and $350$ literal
\texttt{[action~tokens]}), spanning high-resource world languages such as
Spanish through the low-resource frontier. These modifications are integrated
into the main TextArena codebase and released under the project's existing MIT
License, which permits modification and redistribution; the original copyright
and license notice are retained. Because a dropped or renamed placeholder crashes
the runtime and a corrupted action token silently breaks playability, the
requirement is not merely fluency but \emph{verifiable structural and semantic
faithfulness}; with no native reviewer to lean on, verifiability---not
translation---is the binding constraint. Two coordinated pipelines share this
requirement and a common determinism layer, differing only where the language's
resource level forces a different translator and verifier.


\paragraph{A determinism layer shared by both tracks.}
Before any model sees a string, every must-keep span---\texttt{\{placeholder\}},
escaped-brace literal, \texttt{[action token]}, backticked code, and card
label---is replaced by an ordered sentinel and reinserted afterward. Token
preservation thus becomes a property we \emph{enforce and check} (the sentinel
multiset in the output must equal the input's) rather than hope a model
respects. We thus retire token-corruption and placeholder-loss failures by construction
for machine-translation and LLM outputs alike. A separate deterministic
parser-token oracle re-derives each game's action grammar from its environment
code and confirms that every literal keyword and concrete example survives
translation, which \emph{guarantees structural playability independent of prose quality}. The
sentinel form was chosen empirically: of twelve candidates only three survived
machine translation verbatim across all tested scripts (including right-to-left
Arabic), and of those only the CJK corner bracket avoids colliding with the
corpus's own \texttt{[tokens]} and \texttt{\{placeholders\}}.

\paragraph{Higher- and mid-resource languages.}
For languages that strong instruction models read well, translation uses a gateway
serving Llama-3.1-405B and Qwen2.5-72B, and quality is enforced by a layered
pipeline: cheap deterministic filters (script and homoglyph contamination,
literal-keyword translation, brace-literal damage), the parser-token oracle, and a
two-stage semantic gate---a full-corpus sweep by a generation-tier model followed
by confirmation from an independent, more capable reviewer. Across the six-language
batch we audit in detail here---a subset of the released higher/mid-resource
set---the deterministic tiers eliminated the entire class of functional defects
($0$ keyword or example losses across $65\times6$ games), and
the semantic sweep flagged $2.9\%$ of game--language pairs, of which confirmation
kept ten genuine \emph{prose} mistranslations---an inverted objective, a wrong
quantifier, a mislabeled domain term---\emph{all} of them in games outside the
hand-inspected sample---which is why the semantic sweep must be
exhaustive rather than a spot-check. The pipeline was verified against the arabic translations that were manually verified and, the same detector found essentially zero
residual defects in both the reference corpus and the fully automatic output while still catching
real defects in an adversarial generation. While we cannot have guarantees, this strengthens the belief that the process is reliable.

\paragraph{The low-resource frontier.}
Below the band that gateway models read, two assumptions of the first track fail:
no served model translates the language competently, and no author or judge can be
assumed to read it. Translation therefore uses open multilingual machine
translation \citep[NLLB-200;][]{costa2022no}, and verification must be \emph{reader-free}. We localized
$143$ additional low-resource languages this way; the central methodological
result of this track concerns the verifier, and it is cautionary.

\paragraph{Reader-free verification: what fails, and what works.}
The natural reader-free check---translate the candidate, back-translate it to
English with independent models, and compare against the source---\emph{over-flags
severely} at the frontier, because the back-translation of a low-resource language
\emph{into} English is itself unreliable and injects spurious disagreement.
Measured against a direct judge it reported $108$, $162$, and $317$ divergences for
Hausa, Yoruba, and Twi where essentially none are real. A fixed language identifier
(fastText \texttt{lid.176}) is likewise unusable here, misclassifying fluent,
correct Hausa as a different language. The reliable instrument is instead a
\emph{direct} bilingual fidelity judge: a capable instruction model reads the
English source and the candidate together and scores faithfulness, run
\emph{carefully} (one string at a time, with an explicit instruction to check
facts, quantities, and entities) and required to \emph{agree across two model
families}. Validated on the human-known control it reaches $100\%$ sensitivity and
specificity, whereas a batched version of the same judge catches only about
a third of deliberately corrupted translations.

\paragraph{Tiered labeling.}
Every string that the verifier confirms wrong, or whose tokens cannot be restored
safely, is reverted to English, so no known-wrong string ships; the cost is a
measured English-fallback fraction reported per language rather than a silent
error. Languages are tiered by measured meaning fidelity on a sampled audit: those
at or above $85\%$ (median $98\%$) are labeled certified-flagged, the remainder
experimental. We are explicit that this is \emph{machine verification, not native
review}: each shipped language publishes its measured fidelity and target-language
coverage, and a runtime helper warns when a non-certified locale is loaded. The
residual weak tail---languages where even careful two-model judging is uncertain
because the judge's own competence is limited---marks the ceiling of
open-model localization, beyond which native review, not further automation, is the
only lever. Consistent with this, a family-specialist study found only narrow
gains: a purpose-built African translation model improved one of fifteen weak
languages and was worse on the other fourteen, confirming that general machine
translation with careful verification, not specialist translators, is the workhorse
at this scale.

\paragraph{Scale and cost.}
Localization compute is not the bottleneck; verifiability is. The full decode
across the target languages is on the order of tens of GPU-hours, dominated by the
low-resource machine translation; the scarce resource is a verifier trustworthy in
languages no author reads, which the careful two-model direct judge supplies. All
produced locale files and per-language confidence records are released with TextArena; the tooling that reproduces the pipeline is kept on a
separate branch for reproducibility, but we did not ask the TextArena maintainers to merge it.

\paragraph{Language inventory.}
Tab.~\ref{tab:tracka} and Tab.~\ref{tab:trackb} enumerate every localized language with its ISO~639-3 code, dominant script (ISO~15924), the forward translation model, and two independent resource signals: the resource class of  \citet[0 = least-resourced, 5 = most]{joshi2020state} and the number of speakers recorded in Wikidata  \cite[property~P1098][]{vrandecic2014wikidata}, with `--' where a source has no entry. Both tables are ordered from higher- to lower-resource (by Joshi class, then speaker count), so the descent into the long tail is visible top-to-bottom. Table~\ref{tab:tracka} lists Track~A: the 8 core experiment languages (localized with GPT-5.2 and Opus~4.8 under native-speaker review) and 42 higher/mid-resource expansion languages translated with Llama-3.1-405B \cite{grattafiori2024llama3herdmodels} (Qwen2.5-72B \cite{qwen25} for CJK and Southeast-Asian scripts) and verified by an independent, more capable LLM rather than a native reviewer. Tab.~\ref{tab:trackb}--\ref{tab:trackbEnd} list the 143 languages of Track~B, the low-resource frontier. Here the forward translation is produced by the dedicated machine translation (MT) model NLLB-200 \cite{costa2022no} (Chokwe via the Toucan African-language model, \citealp{elmadany2024toucan}); meaning fidelity is then checked by a careful two-model LLM judge---Llama-3.1-405B \cite{grattafiori2024llama3herdmodels} and Qwen2.5-72B \cite{qwen25}, per-leaf concordance---rather than a native reviewer, with MADLAD-400 \cite{kudugunta2023madlad} used alongside NLLB for back-translation calibration and family-specialist MT (IndicTrans2, \citealp{gala2023indictrans2}, Toucan, \citealp{elmadany2024toucan}) trialed on the Indic and African tails, where it gave only narrow gains over NLLB. For Track~B we additionally report the measured meaning-fidelity and target-language coverage percentages that assign each language its tier.

\begin{table*}[p]
\centering
\footnotesize
\setlength{\tabcolsep}{4pt}\renewcommand{\arraystretch}{0.95}
\caption{Track~A languages (higher/mid-resource): 8 core experiment languages plus 42 expansion languages, ordered higher- to lower-resource. \emph{Translator} is the model that produced the localization; \emph{Rev.} distinguishes native human review (\emph{native}, core set) from LLM-only verification (\emph{LLM}, expansion). \emph{Class}: Joshi et al.\ \cite{joshi2020state} resource class; \emph{Speakers}: Wikidata P1098 \cite{vrandecic2014wikidata}.}\label{tab:tracka}
\begin{tabular}{llllcccr}
\toprule
Language & ISO & Script & Translator & Rev. & Tier & Class & Speakers \\
\midrule
Chinese & zho & Hans & GPT-5.2 / Opus & native & A & 5 & 1299.9M \\
English & eng & Latn & GPT-5.2 / Opus & native & A & 5 & 753.4M \\
Spanish & spa & Latn & GPT-5.2 / Opus & native & A & 5 & 485.0M \\
Arabic & ara & Arab & GPT-5.2 / Opus & native & A & 5 & 315.4M \\
French & fra & Latn & GPT-5.2 / Opus & native & A & 5 & 208.2M \\
Japanese & jpn & Jpan & Qwen2.5-72B & LLM & B & 5 & 128.0M \\
German & deu & Latn & GPT-5.2 / Opus & native & A & 5 & 76.5M \\
Hindi & hin & Deva & Llama-3.1-405B & LLM & B & 4 & 341.0M \\
Portuguese & por & Latn & Llama-3.1-405B & LLM & B & 4 & 254.3M \\
Russian & rus & Cyrl & Llama-3.1-405B & LLM & B & 4 & 154.0M \\
Turkish & tur & Latn & Llama-3.1-405B & LLM & B & 4 & 82.2M \\
Korean & kor & Kore & Qwen2.5-72B & LLM & B & 4 & 77.3M \\
Vietnamese & vie & Latn & Qwen2.5-72B & LLM & B & 4 & 76.0M \\
Italian & ita & Latn & Llama-3.1-405B & LLM & B & 4 & 64.8M \\
Persian & fas & Arab & Llama-3.1-405B & LLM & B & 4 & 45.0M \\
Polish & pol & Latn & Llama-3.1-405B & LLM & B & 4 & 39.7M \\
Dutch & nld & Latn & Llama-3.1-405B & LLM & B & 4 & 23.1M \\
Hungarian & hun & Latn & Llama-3.1-405B & LLM & B & 4 & 12.6M \\
Czech & ces & Latn & Llama-3.1-405B & LLM & B & 4 & 10.7M \\
Swedish & swe & Latn & Llama-3.1-405B & LLM & B & 4 & 9.2M \\
Serbian (Cyrillic) & srp & Cyrl & Llama-3.1-405B & LLM & B & 4 & 9.0M \\
Croatian & hrv & Latn & Llama-3.1-405B & LLM & B & 4 & 7.0M \\
Finnish & fin & Latn & Llama-3.1-405B & LLM & B & 4 & 5.4M \\
Catalan & cat & Latn & Llama-3.1-405B & LLM & B & 4 & 4.9M \\
Bengali & ben & Beng & Llama-3.1-405B & LLM & B & 3 & 300.0M \\
Indonesian & ind & Latn & Qwen2.5-72B & LLM & B & 3 & 199.0M \\
Filipino & fil & Latn & Llama-3.1-405B & LLM & B & 3 & 90.0M \\
Malay & msa & Latn & GPT-5.2 / Opus & native & A & 3 & 77.0M \\
Tamil & tam & Taml & Llama-3.1-405B & LLM & B & 3 & 75.0M \\
Urdu & urd & Arab & Llama-3.1-405B & LLM & B & 3 & 68.6M \\
Ukrainian & ukr & Cyrl & Llama-3.1-405B & LLM & B & 3 & 26.9M \\
Romanian & ron & Latn & Llama-3.1-405B & LLM & B & 3 & 24.3M \\
Thai & tha & Thai & Qwen2.5-72B & LLM & B & 3 & 20.7M \\
Greek & ell & Grek & Llama-3.1-405B & LLM & B & 3 & 15.0M \\
Afrikaans & afr & Latn & Llama-3.1-405B & LLM & B & 3 & 10.3M \\
Hebrew & heb & Hebr & GPT-5.2 / Opus & native & A & 3 & 9.3M \\
Bulgarian & bul & Cyrl & Llama-3.1-405B & LLM & B & 3 & 9.0M \\
Danish & dan & Latn & Llama-3.1-405B & LLM & B & 3 & 6.0M \\
Slovak & slk & Latn & Llama-3.1-405B & LLM & B & 3 & 6.0M \\
Lithuanian & lit & Latn & Llama-3.1-405B & LLM & B & 3 & 4.0M \\
Galician & glg & Latn & Llama-3.1-405B & LLM & B & 3 & 2.4M \\
Slovenian & slv & Latn & Llama-3.1-405B & LLM & B & 3 & 2.4M \\
Latvian & lav & Latn & Llama-3.1-405B & LLM & B & 3 & 1.5M \\
Estonian & est & Latn & Llama-3.1-405B & LLM & B & 3 & 1.3M \\
Swahili & swa & Latn & Llama-3.1-405B & LLM & B & 2 & 15.4M \\
Icelandic & isl & Latn & Llama-3.1-405B & LLM & B & 2 & 321k \\
Azerbaijani & aze & Latn & Llama-3.1-405B & LLM & B & 1 & 23.0M \\
Albanian & sqi & Latn & Llama-3.1-405B & LLM & B & 1 & 6.2M \\
Norwegian Bokmål & nob & Latn & Llama-3.1-405B & LLM & B & 1 & 4.0M \\
Macedonian & mkd & Cyrl & Llama-3.1-405B & LLM & B & 1 & 2.0M \\
\bottomrule
\end{tabular}
\end{table*}

\begin{table*}[p]
\centering
\footnotesize
\setlength{\tabcolsep}{4pt}\renewcommand{\arraystretch}{0.95}
\caption{Track~B languages (low-resource tier), ordered higher- to lower-resource. \emph{Translator} is the forward MT model: NLLB-200 \cite{costa2022no} for all except Chokwe (Toucan \cite{elmadany2024toucan}); meaning fidelity was then verified by a two-model LLM judge (Llama-3.1-405B \cite{grattafiori2024llama3herdmodels} + Qwen2.5-72B \cite{qwen25}), not by a native reviewer. \emph{Tier}: C = certified-flagged (fidelity $\geq$85\%), E = experimental. \emph{Class}: Joshi et al.\ \cite{joshi2020state}; \emph{Speakers}: Wikidata P1098 \cite{vrandecic2014wikidata}. \emph{Fid.}: measured meaning-fidelity \%; \emph{Cov.}: target-language coverage \%.}\label{tab:trackb}
\begin{tabular}{llllccrcc}
\toprule
Language & ISO & Script & Translator & Tier & Class & Speakers & Fid. & Cov. \\
\midrule
North Levantine Arabic & apc & Arab & NLLB-200 & C & 5 & 44.0M & 98 & 91 \\
Moroccan Arabic & ary & Arab & NLLB-200 & C & 5 & 27.5M & 100 & 91 \\
Mesopotamian Arabic & acm & Latn & NLLB-200 & C & 5 & 15.7M & 95 & 92 \\
South Levantine Arabic & ajp & Latn & NLLB-200 & C & 5 & 11.6M & 98 & 92 \\
Tunisian Arabic & aeb & Arab & NLLB-200 & C & 5 & 11.6M & 90 & 90 \\
Taizzi-Adeni Arabic & acq & Latn & NLLB-200 & C & 5 & 10.5M & 95 & 92 \\
Najdi Arabic & ars & Arab & NLLB-200 & C & 5 & -- & 98 & 93 \\
Basque & eus & Latn & NLLB-200 & C & 4 & 750k & 90 & 86 \\
Egyptian Arabic & arz & Arab & NLLB-200 & C & 3 & 64.6M & 100 & 92 \\
Uzbek & uzb & Latn & NLLB-200 & C & 3 & 27.0M & 100 & 95 \\
Cebuano & ceb & Latn & NLLB-200 & C & 3 & 15.9M & 98 & 91 \\
Kazakh & kaz & Cyrl & NLLB-200 & C & 3 & 12.9M & 98 & 93 \\
Belarusian & bel & Cyrl & NLLB-200 & C & 3 & 7.6M & 98 & 96 \\
Georgian & kat & Geor & NLLB-200 & C & 3 & 3.7M & 95 & 91 \\
Punjabi & pan & Guru & NLLB-200 & C & 2 & 125.0M & 100 & 97 \\
Marathi & mar & Deva & NLLB-200 & C & 2 & 83.1M & 100 & 97 \\
Hausa & hau & Latn & NLLB-200 & C & 2 & 43.9M & 98 & 95 \\
Yoruba & yor & Latn & NLLB-200 & C & 2 & 37.8M & 100 & 93 \\
Amharic & amh & Ethi & NLLB-200 & C & 2 & 21.9M & 98 & 85 \\
Zulu & zul & Latn & NLLB-200 & C & 2 & 12.1M & 98 & 95 \\
Xhosa & xho & Latn & NLLB-200 & C & 2 & 8.2M & 100 & 86 \\
Tigrinya & tir & Ethi & NLLB-200 & C & 2 & 7.5M & 95 & 78 \\
Lao & lao & Laoo & NLLB-200 & C & 2 & 5.2M & 100 & 95 \\
Tswana & tsn & Latn & NLLB-200 & C & 2 & 4.5M & 92 & 86 \\
Wolof & wol & Latn & NLLB-200 & C & 2 & 3.7M & 95 & 70 \\
Maltese & mlt & Latn & NLLB-200 & C & 2 & 570k & 100 & 86 \\
Irish & gle & Latn & NLLB-200 & C & 2 & 141k & 100 & 89 \\
Sanskrit & san & Deva & NLLB-200 & C & 2 & 50k & 98 & 86 \\
Telugu & tel & Telu & NLLB-200 & C & 1 & 82.0M & 98 & 95 \\
Javanese & jav & Latn & NLLB-200 & C & 1 & 68.3M & 100 & 93 \\
Gujarati & guj & Gujr & NLLB-200 & C & 1 & 56.4M & 100 & 96 \\
Bhojpuri & bho & Deva & NLLB-200 & C & 1 & 52.2M & 98 & 88 \\
Kannada & kan & Knda & NLLB-200 & C & 1 & 43.6M & 100 & 95 \\
Pashto & pus & Arab & NLLB-200 & C & 1 & 39.0M & 98 & 95 \\
Malayalam & mal & Mlym & NLLB-200 & C & 1 & 37.1M & 100 & 95 \\
Odia & ori & Orya & NLLB-200 & C & 1 & 34.5M & 100 & 94 \\
Maithili & mai & Deva & NLLB-200 & C & 1 & 33.9M & 100 & 91 \\
Burmese & mya & Mymr & NLLB-200 & C & 1 & 32.9M & 95 & 94 \\
Sundanese & sun & Latn & NLLB-200 & C & 1 & 32.4M & 95 & 94 \\
Igbo & ibo & Latn & NLLB-200 & C & 1 & 27.0M & 98 & 96 \\
Sindhi & snd & Arab & NLLB-200 & C & 1 & 24.6M & 100 & 96 \\
Lingala & lin & Latn & NLLB-200 & E & 1 & 20.0M & 80 & 85 \\
Malagasy & mlg & Latn & NLLB-200 & C & 1 & 18.0M & 95 & 94 \\
Khmer & khm & Khmr & NLLB-200 & C & 1 & 16.6M & 95 & 94 \\
Somali & som & Latn & NLLB-200 & C & 1 & 16.2M & 92 & 95 \\
Turkmen & tuk & Latn & NLLB-200 & C & 1 & 16.0M & 100 & 86 \\
Nepali & nep & Deva & NLLB-200 & C & 1 & 15.8M & 95 & 95 \\
Assamese & asm & Beng & NLLB-200 & C & 1 & 15.3M & 100 & 95 \\
\bottomrule
\end{tabular}
\end{table*}

\begin{table*}[p]
\centering
\footnotesize
\setlength{\tabcolsep}{4pt}\renewcommand{\arraystretch}{0.95}
\caption{Track~B languages (low-resource tier), \emph{continued}.}
\begin{tabular}{llllccrcc}
\toprule
Language & ISO & Script & Translator & Tier & Class & Speakers & Fid. & Cov. \\
\midrule
Northern Kurdish & kmr & Latn & NLLB-200 & C & 1 & 14.6M & 98 & 87 \\
Tajik & tgk & Cyrl & NLLB-200 & C & 1 & 14.0M & 100 & 90 \\
South Azerbaijani & azb & Latn & NLLB-200 & C & 1 & 13.8M & 95 & 82 \\
Tsonga & tso & Latn & NLLB-200 & C & 1 & 13.0M & 95 & 88 \\
Kinyarwanda & kin & Latn & NLLB-200 & C & 1 & 12.1M & 92 & 86 \\
Nyanja & nya & Latn & NLLB-200 & E & 1 & 12.0M & 82 & 90 \\
Akan & aka & Latn & NLLB-200 & C & 1 & 11.0M & 98 & 74 \\
Uyghur & uig & Arab & NLLB-200 & C & 1 & 10.4M & 92 & 82 \\
Ilocano & ilo & Latn & NLLB-200 & C & 1 & 9.1M & 100 & 89 \\
Shona & sna & Latn & NLLB-200 & C & 1 & 8.3M & 98 & 90 \\
Central Kurdish & ckb & Arab & NLLB-200 & C & 1 & 7.2M & 100 & 86 \\
Santali & sat & Olck & NLLB-200 & E & 1 & 7.2M & 80 & 75 \\
Tumbuka & tum & Latn & NLLB-200 & E & 1 & 7.0M & 72 & 82 \\
Kashmiri & kas & Arab & NLLB-200 & C & 1 & 6.9M & 88 & 84 \\
Armenian & hye & Armn & NLLB-200 & C & 1 & 6.7M & 95 & 95 \\
Kikuyu & kik & Latn & NLLB-200 & E & 1 & 6.6M & 78 & 79 \\
Southern Sotho & sot & Latn & NLLB-200 & C & 1 & 6.0M & 85 & 89 \\
Kabyle & kab & Latn & NLLB-200 & C & 1 & 5.6M & 90 & 80 \\
Minangkabau & min & Latn & NLLB-200 & C & 1 & 5.5M & 100 & 90 \\
Mongolian & mon & Cyrl & NLLB-200 & C & 1 & 5.2M & 98 & 90 \\
Tatar & tat & Cyrl & NLLB-200 & C & 1 & 5.2M & 100 & 86 \\
Buginese & bug & Latn & NLLB-200 & E & 1 & 5.0M & 80 & 86 \\
Kikongo & kon & Latn & NLLB-200 & E & 1 & 5.0M & 80 & 76 \\
Sicilian & scn & Latn & NLLB-200 & C & 1 & 4.7M & 100 & 92 \\
Sango & sag & Latn & NLLB-200 & E & 1 & 4.6M & 82 & 77 \\
Kyrgyz & kir & Cyrl & NLLB-200 & C & 1 & 4.6M & 98 & 94 \\
Guarani & grn & Latn & NLLB-200 & C & 1 & 4.5M & 90 & 84 \\
Norwegian Nynorsk & nno & Latn & NLLB-200 & C & 1 & 4.3M & 92 & 89 \\
Bambara & bam & Latn & NLLB-200 & E & 1 & 4.2M & 80 & 79 \\
Ganda & lug & Latn & NLLB-200 & C & 1 & 4.1M & 92 & 84 \\
Northern Sotho & nso & Latn & NLLB-200 & C & 1 & 4.1M & 88 & 85 \\
Tok Pisin & tpi & Latn & NLLB-200 & C & 1 & 4.0M & 90 & 84 \\
Lombard & lmo & Latn & NLLB-200 & C & 1 & 3.9M & 100 & 88 \\
Acehnese & ace & Latn & NLLB-200 & C & 1 & 3.5M & 100 & 92 \\
Banjar & bjn & Latn & NLLB-200 & C & 1 & 3.5M & 100 & 92 \\
Waray & war & Latn & NLLB-200 & C & 1 & 3.1M & 98 & 87 \\
Ewe & ewe & Latn & NLLB-200 & C & 1 & 3.0M & 90 & 79 \\
Twi & twi & Latn & NLLB-200 & C & 1 & 3.0M & 92 & 76 \\
Swati & ssw & Latn & NLLB-200 & C & 1 & 2.0M & 92 & 87 \\
Esperanto & epo & Latn & NLLB-200 & C & 1 & 2.0M & 100 & 94 \\
Venetian & vec & Latn & NLLB-200 & C & 1 & 2.0M & 100 & 92 \\
Limburgish & lim & Latn & NLLB-200 & C & 1 & 1.6M & 95 & 91 \\
Sardinian & srd & Latn & NLLB-200 & C & 1 & 1.3M & 100 & 88 \\
Bashkir & bak & Cyrl & NLLB-200 & C & 1 & 1.2M & 100 & 82 \\
Standard Tibetan & bod & Tibt & NLLB-200 & C & 1 & 1.2M & 95 & 75 \\
Pangasinan & pag & Latn & NLLB-200 & C & 1 & 1.1M & 88 & 86 \\
Kabiye & kbp & Latn & NLLB-200 & E & 1 & 1.0M & 78 & 64 \\
Ayacucho Quechua & quy & Latn & NLLB-200 & C & 1 & 918k & 90 & 79 \\
\bottomrule
\end{tabular}
\end{table*}

\begin{table*}[p]
\centering
\footnotesize
\setlength{\tabcolsep}{4pt}\renewcommand{\arraystretch}{0.95}
\caption{Track~B languages (low-resource tier), \emph{continued}.}
\label{tab:trackbEnd}
\begin{tabular}{llllccrcc}
\toprule
Language & ISO & Script & Translator & Tier & Class & Speakers & Fid. & Cov. \\
\midrule
Welsh & cym & Latn & NLLB-200 & C & 1 & 724k & 95 & 93 \\
Crimean Tatar & crh & Cyrl & NLLB-200 & C & 1 & 553k & 100 & 90 \\
Occitan & oci & Latn & NLLB-200 & C & 1 & 542k & 100 & 91 \\
Ligurian & lij & Latn & NLLB-200 & C & 1 & 500k & 100 & 87 \\
Silesian & szl & Latn & NLLB-200 & C & 1 & 458k & 98 & 91 \\
Asturian & ast & Latn & NLLB-200 & C & 1 & 450k & 98 & 84 \\
Samoan & smo & Latn & NLLB-200 & C & 1 & 416k & 98 & 89 \\
Luxembourgish & ltz & Latn & NLLB-200 & C & 1 & 391k & 100 & 90 \\
Fijian & fij & Latn & NLLB-200 & C & 1 & 341k & 90 & 84 \\
Friulian & fur & Latn & NLLB-200 & C & 1 & 300k & 98 & 88 \\
Dzongkha & dzo & Tibt & NLLB-200 & C & 1 & 237k & 98 & 75 \\
Maori & mri & Latn & NLLB-200 & C & 1 & 214k & 100 & 90 \\
Latgalian & ltg & Latn & NLLB-200 & C & 1 & 200k & 95 & 87 \\
Faroese & fao & Latn & NLLB-200 & C & 1 & 69k & 98 & 91 \\
Scottish Gaelic & gla & Latn & NLLB-200 & C & 1 & 60k & 98 & 88 \\
Central Aymara & ayr & Latn & NLLB-200 & C & 1 & -- & 92 & 79 \\
Eastern Yiddish & ydd & Hebr & NLLB-200 & C & 1 & -- & 98 & 88 \\
Southwestern Dinka & dik & Latn & NLLB-200 & C & 1 & -- & 85 & 78 \\
West Central Oromo & gaz & Latn & NLLB-200 & E & 1 & -- & 75 & 79 \\
Awadhi & awa & Deva & NLLB-200 & C & 0 & 22.0M & 98 & 89 \\
Magahi & mag & Deva & NLLB-200 & C & 0 & 20.7M & 100 & 88 \\
Central Atlas Tamazight & tzm & Latn & NLLB-200 & C & 0 & 17.0M & 98 & 76 \\
Sinhala & sin & Sinh & NLLB-200 & C & 0 & 15.3M & 95 & 94 \\
Nigerian Fulfulde & fuv & Latn & NLLB-200 & C & 0 & 14.5M & 92 & 81 \\
Rundi & run & Latn & NLLB-200 & E & 0 & 10.8M & 80 & 86 \\
Haitian Creole & hat & Latn & NLLB-200 & C & 0 & 9.6M & 98 & 93 \\
Central Kanuri & knc & Latn & NLLB-200 & C & 0 & 9.3M & 92 & 76 \\
Luba-Kasai & lua & Latn & NLLB-200 & E & 0 & 6.3M & 65 & 82 \\
Umbundu & umb & Latn & NLLB-200 & E & 0 & 6.0M & 72 & 63 \\
Balinese & ban & Latn & NLLB-200 & C & 0 & 4.0M & 98 & 91 \\
Kamba & kam & Latn & NLLB-200 & E & 0 & 3.9M & 80 & 57 \\
Bemba & bem & Latn & NLLB-200 & C & 0 & 3.6M & 85 & 78 \\
Shan & shn & Mymr & NLLB-200 & C & 0 & 3.0M & 98 & 76 \\
Dyula & dyu & Latn & NLLB-200 & E & 0 & 2.7M & 82 & 68 \\
Fon & fon & Latn & NLLB-200 & C & 0 & 1.9M & 90 & 74 \\
Jingpho & kac & Latn & NLLB-200 & C & 0 & 940k & 98 & 71 \\
Nuer & nus & Latn & NLLB-200 & C & 0 & 900k & 88 & 79 \\
Mizo & lus & Latn & NLLB-200 & C & 0 & 500k & 98 & 78 \\
Tamasheq & taq & Latn & NLLB-200 & C & 0 & 500k & 98 & 71 \\
Chhattisgarhi & hne & Deva & NLLB-200 & C & -- & 16.3M & 100 & 88 \\
Mossi & mos & Latn & NLLB-200 & C & -- & 7.6M & 92 & 71 \\
Luo & luo & Latn & NLLB-200 & C & -- & 3.0M & 92 & 84 \\
Meitei & mni & Beng & NLLB-200 & E & -- & 1.5M & 78 & 69 \\
Kabuverdianu & kea & Latn & NLLB-200 & C & -- & 871k & 95 & 87 \\
Papiamento & pap & Latn & NLLB-200 & C & -- & 321k & 100 & 92 \\
Chokwe & cjk & Latn & Toucan & E & -- & -- & 63 & 63 \\
Kimbundu & kmb & Latn & NLLB-200 & E & -- & -- & 70 & 63 \\
\bottomrule
\end{tabular}
\end{table*}

\clearpage

\section{Per-game language strength}

\begin{figure*}[!t]
\centering

\begin{subfigure}[t]{0.95\textwidth}
    \centering
    \includegraphics[width=\linewidth]{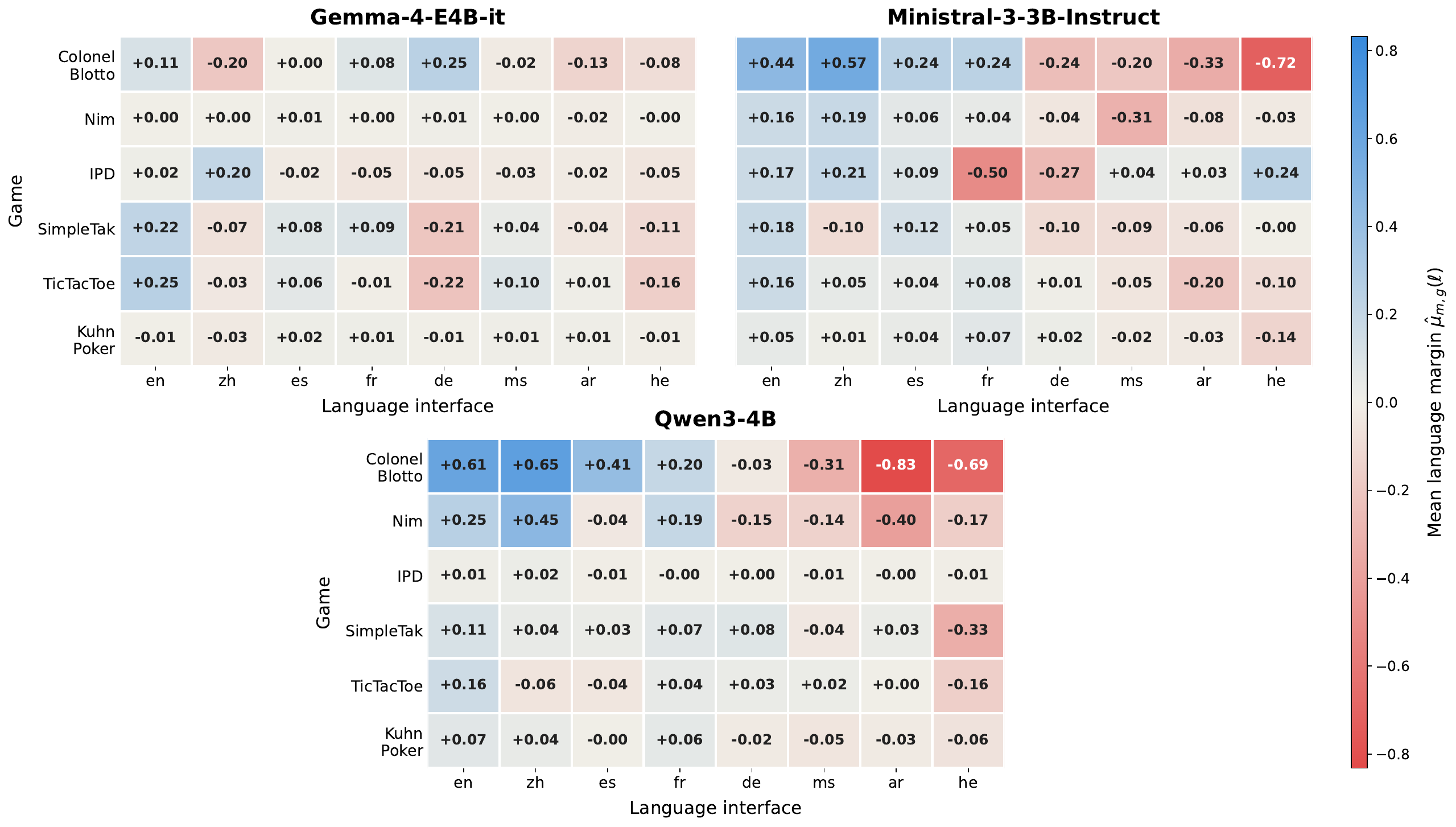}
\end{subfigure}

\caption{
Per-game language strength profiles for Gemma-4-E4B-it, Ministral-3-3B-Instruct, and Qwen3-4B. Each cell reports the mean language margin $\mu_{m,g}(A)$ for model $m$, game $g$, and language interface $A$, obtained by averaging the role-pooled pairwise margins of $A$ against all other evaluated languages. Positive values indicate stronger average performance through language $A$, while negative values indicate weaker average performance.
}
\label{fig:per_game_language_strength_profiles}

\end{figure*}

Fig.~\ref{fig:per_game_language_strength_profiles} decomposes each model's language profile by game, reporting the mean language win--loss margin $\mu_{m,g}(A)$ for every model--game--language. $\Delta_{m,g}(A,B) = -\Delta_{m,g}(B,A)$, each row sums to zero by construction; cells therefore measure the \emph{relative} strength of an interface within a model--game pair rather than absolute playing quality.

\section{Detailed Skill Analyses}
In the following sections, we expand our findings on Sec.~\ref{sec:skill_differences}. Unless otherwise specified, Gemma, Qwen, and Ministral refer to their 4B-sized models.
\label{app:skill_analyses}
\subsection{Differences in Spatial Reasoning}
\label{sec:spatialreason}

Since text representations serialize content row by row, LLMs may track rows more easily than columns or diagonals. We therefore analyze losses in TicTacToe and SimpleTak, both of which present a 2D board in the observation \(o_t\) (see App.~\ref{app:ttt} and App.~\ref{app:st}). Because losses often arise when a model fails to detect or respond to a positional threat, the rows, columns, or diagonals along which it loses provide a direct view of its spatial reasoning limitations.

\paragraph{TicTacToe}

\begin{table}[h]
\centering
\footnotesize
\setlength{\tabcolsep}{4.0pt}
\renewcommand{\arraystretch}{1.02}

\begin{tabular*}{\columnwidth}{
    @{\extracolsep{\fill}}
    l
    rrr
    @{}
}
\toprule
\textbf{Language}
& \textbf{Row}
& \textbf{Column}
& \textbf{Diagonal} \\
\midrule

\multicolumn{4}{@{}l}{\textbf{Gemma-4-E4B-it}} \\
\addlinespace[1pt]
English & 28.5\% & 34.8\% & 36.7\% \\
Chinese & 17.5\% & 28.0\% & 54.5\% \\
Spanish & 18.5\% & 34.4\% & 47.0\% \\
French  & 22.6\% & 35.6\% & 41.7\% \\
German  & 19.8\% & 37.5\% & 42.8\% \\
Hebrew  & 21.1\% & 35.6\% & 43.3\% \\
Arabic  & 20.3\% & 34.4\% & 45.3\% \\
Malay   & 20.7\% & 29.0\% & 50.3\% \\

\midrule
\multicolumn{4}{@{}l}{\textbf{Qwen3-4B}} \\
\addlinespace[1pt]
English & 45.3\% & 23.8\% & 30.9\% \\
Chinese & 44.0\% & 22.9\% & 33.1\% \\
Spanish & 49.8\% & 19.9\% & 30.3\% \\
French  & 48.5\% & 22.6\% & 28.8\% \\
German  & 48.0\% & 24.5\% & 27.4\% \\
Hebrew  & 39.7\% & 31.3\% & 29.0\% \\
Arabic  & 45.7\% & 23.8\% & 30.5\% \\
Malay   & 46.8\% & 21.8\% & 31.3\% \\

\midrule
\multicolumn{4}{@{}l}{\textbf{Ministral3-3B}} \\
\addlinespace[1pt]
English & 42.1\% & 35.3\% & 22.6\% \\
Chinese & 35.0\% & 33.4\% & 31.7\% \\
Spanish & 39.2\% & 31.7\% & 29.2\% \\
French  & 37.9\% & 34.5\% & 27.6\% \\
German  & 37.3\% & 33.9\% & 28.8\% \\
Hebrew  & 33.4\% & 32.8\% & 33.9\% \\
Arabic  & 35.0\% & 33.7\% & 31.4\% \\
Malay   & 33.3\% & 33.8\% & 32.9\% \\
\bottomrule
\end{tabular*}

\caption{
Distribution of defeat types in \textsc{TicTacToe} by language and
model. Each value is the percentage of defeats in which the opponent completed a row, column, or diagonal. These distributions indicate which spatial relationships each model fails to track under different language interfaces.
}
\label{tab:tictactoeLossType}
\end{table}

When observing Gemma-4-E4B-it~\cite{gemmateam2026gemma4technicalreport} and Ministral3-3B~\cite{liu2026ministral3} in TicTacToe, we find that non-English interfaces, and more so low-resource or non-Latin-script languages, show a skew towards column and diagonal losses compared to the English interface. We attribute this finding to the LLMs learning a better, more robust spatial understanding across the different spatial axes. Qwen3-4B~\cite{qwen3technicalreport}, however, shows a more stable defeat pattern across languages, with the exception of the Hebrew interface, suggesting a more balanced or generalized capability of spatial understanding.

For Gemma, English exhibits a relatively balanced distribution of losses across rows (28.5\%), columns (34.8\%), and diagonals (36.7\%), implying a generalized spatial understanding capable of handling multi-dimensional threats. In contrast Arabic and Hebrew loss patterns skew away from rows, making them more susceptible to column and diagonal threats. Arabic's loss pattern is rows (20.3\%), columns (34.4\%), diagonals (45.3\%), and Hebrew follows similarly. This quantitative analysis, suggesting a spatial vulnerability, is strongly corroborated by our qualitative analysis, where Arabic and Hebrew game trajectories showed frequent "hallucination" or mislabeling of series of cells as columns or diagonals.

\paragraph{Simple Tak}
\begin{table}[!ht]
\centering
\footnotesize
\begin{tabular*}{\columnwidth}{
    @{\extracolsep{\fill}}
    l cc @{}
}
\toprule
\textbf{Language} & \textbf{Row} & \textbf{Column} \\
\midrule
English  & 53.2\% & 46.8\% \\
Chinese  & 52.3\% & 47.7\% \\
Spanish  & 51.5\% & 48.5\% \\
French   & 46.0\% & 54.0\% \\
German   & 39.7\% & 60.3\% \\
Hebrew   & 42.8\% & 57.2\% \\
Arabic   & 42.8\% & 57.2\% \\
Malay    & 41.9\% & 58.1\% \\
\bottomrule
\end{tabular*}
\caption{Distribution of loss types in \textsc{SimpleTak} by language for Gemma4-E4B-it: percentage of losses where the opponent completed a row or column.}
\label{tab:simpletakLossType}
\end{table}
Simple Tak's winning lines can include both horizontal and vertical losses intertwined, which means analysis of its loss distribution should seemingly be more complex. However, in practice it seems Gemma-4-E4B-it favors straight lines (over 90\% of wins), which permits associating horizontal and vertical lines with different aspects of spatial understanding as we did for TicTacToe. We present a detailed per-language breakdown of loss type distribution in Tab.~\ref{tab:simpletakLossType}. Similarly to TicTacToe, some non-Latin and low-resource languages show a skew toward column losses compared to English. Specifically, while English loss distribution is rows 53.2\% and columns 46.8\%, Arabic and Hebrew show a skew towards columns with a loss distribution of rows 42.8\% and columns 57.2\%.

\subsection{Differences in Strategy}
\label{sec:strategy}

\begin{table}[h]
\centering
\scriptsize
\setlength{\tabcolsep}{2.2pt}
\renewcommand{\arraystretch}{0.98}

\begin{tabular*}{\columnwidth}{
    @{\extracolsep{\fill}}
    l
    rrrrrr
    @{}
}
\toprule
\textbf{Lang.}
& \textbf{Inv.}
& \textbf{Bluff$_J$}
& \textbf{Bet$_Q$}
& \textbf{Value$_K$}
& \textbf{Call$_K$}
& \textbf{Fold$_K$} \\
\midrule

\multicolumn{7}{@{}l}{\textbf{Gemma-4-E4B-it}} \\
\addlinespace[1pt]
English
& 2.5\% & 1.3\% & 42.9\% & 92.2\% & 90.7\% & 2.0\% \\
Chinese
& 2.4\% & 2.4\% & 57.0\% & 92.7\% & 93.7\% & 0.5\% \\
Spanish
& 0.7\% & 0.9\% & 60.3\% & 98.2\% & 98.3\% & 0.1\% \\
French
& 0.5\% & 1.3\% & 59.0\% & 98.3\% & 99.3\% & 0.3\% \\
German
& 0.9\% & 1.9\% & 62.3\% & 96.8\% & 97.8\% & 0.4\% \\
Hebrew
& 1.2\% & 2.4\% & 63.3\% & 96.0\% & 97.7\% & 0.0\% \\
Arabic
& 2.0\% & 2.3\% & 54.1\% & 94.5\% & 96.5\% & 0.4\% \\
Malay
& 1.1\% & 0.8\% & 41.6\% & 95.6\% & 97.8\% & 0.2\% \\
\cmidrule(lr){1-7}
\textit{Mean}
& 1.4\% & 1.7\% & 55.1\% & 95.5\% & 96.5\% & 0.5\% \\
\textit{SD}
& 0.8\% & 0.7\% & 8.4\% & 2.3\% & 2.9\% & 0.6\% \\

\midrule
\multicolumn{7}{@{}l}{\textbf{Qwen3-4B}} \\
\addlinespace[1pt]
English
& 1.5\% & 23.3\% & 80.1\% & 94.7\% & 99.2\% & 0.2\% \\
Chinese
& 1.8\% & 45.0\% & 81.2\% & 94.6\% & 98.6\% & 1.0\% \\
Spanish
& 2.1\% & 22.5\% & 69.3\% & 86.0\% & 97.9\% & 1.9\% \\
French
& 1.5\% & 26.2\% & 82.9\% & 95.2\% & 98.0\% & 2.0\% \\
German
& 5.7\% & 33.7\% & 71.9\% & 86.4\% & 93.5\% & 3.5\% \\
Hebrew
& 7.0\% & 44.4\% & 58.3\% & 75.3\% & 91.5\% & 6.1\% \\
Arabic
& 3.5\% & 47.4\% & 68.2\% & 84.9\% & 94.5\% & 5.0\% \\
Malay
& 0.8\% & 33.1\% & 61.4\% & 87.1\% & 94.3\% & 5.7\% \\
\cmidrule(lr){1-7}
\textit{Mean}
& 3.0\% & 34.5\% & 71.7\% & 88.0\% & 95.9\% & 3.2\% \\
\textit{SD}
& 2.2\% & 10.1\% & 9.2\% & 6.7\% & 2.8\% & 2.2\% \\

\midrule
\multicolumn{7}{@{}l}{\textbf{Ministral3-3B}} \\
\addlinespace[1pt]
English
& 9.3\% & 47.7\% & 66.6\% & 85.6\% & 80.6\% & 16.3\% \\
Chinese
& 8.8\% & 40.8\% & 59.3\% & 77.8\% & 75.7\% & 16.6\% \\
Spanish
& 10.3\% & 48.2\% & 58.4\% & 77.3\% & 76.8\% & 11.7\% \\
French
& 10.2\% & 36.8\% & 60.1\% & 82.4\% & 76.9\% & 10.4\% \\
German
& 9.6\% & 52.6\% & 60.0\% & 76.1\% & 70.1\% & 18.9\% \\
Hebrew
& 20.9\% & 41.4\% & 46.4\% & 55.1\% & 66.7\% & 23.7\% \\
Arabic
& 13.7\% & 42.0\% & 49.6\% & 57.0\% & 71.8\% & 18.9\% \\
Malay
& 15.6\% & 46.0\% & 52.8\% & 63.4\% & 76.9\% & 12.6\% \\
\cmidrule(lr){1-7}
\textit{Mean}
& 12.3\% & 44.4\% & 56.7\% & 71.8\% & 74.4\% & 16.1\% \\
\textit{SD}
& 4.2\% & 5.1\% & 6.6\% & 11.7\% & 4.5\% & 4.4\% \\
\bottomrule
\end{tabular*}

\caption{
Card-conditioned behavior in \textsc{KuhnPoker} across languages and
models. Inv.\ is the fraction of invalid actions. Bluff$_J$, Bet$_Q$,
and Value$_K$ denote the probabilities of betting with $J$, $Q$, and
$K$, respectively, when \texttt{[check]} and \texttt{[bet]} are
available. Call$_K$ and Fold$_K$ denote the probabilities of calling
and folding with $K$ when facing a bet. Higher Value$_K$ and Call$_K$
indicate more reliable play with the strongest card, whereas a high
Fold$_K$ indicates a severe strategic error. Mean and SD are computed
across languages.
}
\label{tab:kuhnpoker_card_conditioned_behavior}
\end{table}

\paragraph{Kuhn Poker}


Kuhn Poker is a three-card imperfect-information game in which each player receives \(J\), \(Q\), or \(K\) and chooses whether to bet, check, call, or fold; full rules are provided in App.~\ref{app:games}. Its actions are readily interpretable from the private card: betting with \(J\) is a bluff, betting with \(K\) seeks additional payoff from weaker hands, and folding \(K\) when facing a bet is a severe strategic error. Tab.~\ref{tab:kuhnpoker_card_conditioned_behavior} shows that Gemma-4-E4B-it responds most consistently to these distinctions across languages, satisfying \(P(\mathrm{bet}\mid J) < P(\mathrm{bet}\mid Q) < P(\mathrm{bet}\mid K)\) throughout. It bets with \(K\) in at least \(92.2\%\) of eligible decisions, calls with \(K\) more than \(90\%\) of the time, and folds it in at most \(2.0\%\), while bluffing with \(J\) in at most \(2.4\%\). Gemma therefore reliably distinguishes weak from strong private cards, although its near-zero bluffing rate reflects a conservative rather than necessarily optimal policy.

Gemma's largest cross-language variation occurs for the intermediate card \(Q\), whose betting rate ranges from \(41.6\%\) in Malay to \(63.3\%\) in Hebrew; its responses to clearly weak or strong cards are much more stable. English is also not uniformly strongest, yielding Gemma's lowest overall betting rate, lowest Call$_K$ rate, and highest BadFold$_K$ rate. Qwen3-4B is more aggressive and more language-sensitive, with larger shifts in Bluff$_J$, ValueBet$_K$, and BadFold$_K$ across languages. Ministral-3-3B-Instruct is less reliable overall, showing weaker separation between card strengths, higher invalid-action rates, and BadFold$_K$ rates between \(10.4\%\) and \(23.7\%\). Overall, Gemma is the most language-stable in strategically clear states, whereas Qwen and Ministral exhibit larger language-conditioned changes in both strategy and execution. 

\subsection{Differences in pre-existing knowledge}
\label{sec:knowledge}


\begin{table}[h]
\centering
\footnotesize
\setlength{\tabcolsep}{4.0pt}
\renewcommand{\arraystretch}{1.02}

\begin{tabular*}{\columnwidth}{
    @{\extracolsep{\fill}}
    l
    rrr
    @{}
}
\toprule
\textbf{Language}
& \textbf{Win \%}
& \textbf{Opt. Strategy Mentions}
& \textbf{Opt. Move} \\
\midrule

\multicolumn{4}{@{}l}{\textbf{Gemma-4-E4B-it}} \\
\addlinespace[1pt]
English & 50.2\% & 43,927  & 99.9\% \\
Chinese & 50.1\% & 32,207  & 99.8\% \\
Spanish & 50.3\% & 85,103  & 99.7\% \\
French  & 50.1\% & 71,607  & 99.2\% \\
German  & 50.3\% & 54,801  & 99.9\% \\
Hebrew  & 49.8\% & 94,951  & 99.3\% \\
Arabic  & 49.3\% & 139,797 & 99.4\% \\
Malay   & 50.0\% & 53,813  & 99.5\% \\

\midrule
\multicolumn{4}{@{}l}{\textbf{Qwen3-4B}} \\
\addlinespace[1pt]
English & 61.0\% & 150,005 & 80.8\% \\
Chinese & 69.8\% & 85,605  & 74.3\% \\
Spanish & 48.4\% & 103,517 & 38.5\% \\
French  & 58.3\% & 159,675 & 24.6\% \\
German  & 43.6\% & 100,722 & 34.1\% \\
Hebrew  & 42.7\% & 20,002  & 4.0\%  \\
Arabic  & 32.4\% & 24,814  & 10.5\% \\
Malay   & 43.8\% & 70,660  & 26.7\% \\

\midrule
\multicolumn{4}{@{}l}{\textbf{Ministral3-3B}} \\
\addlinespace[1pt]
English & 57.3\% & 538,093 & 34.6\% \\
Chinese & 58.1\% & 537,476 & 20.8\% \\
Spanish & 52.7\% & 629,253 & 22.5\% \\
French  & 52.1\% & 812,283 & 31.3\% \\
German  & 48.2\% & 400,286 & 19.7\% \\
Hebrew  & 48.9\% & 2,753   & 10.8\% \\
Arabic  & 46.2\% & 14,610  & 8.0\%  \\
Malay   & 36.6\% & 236,289 & 15.6\% \\
\bottomrule
\end{tabular*}

\caption{
Cross-lingual \textsc{Nim} performance across three models. We report the overall win rate, the number of optimal strategy mentions, and the \% of successful first move optimal plays (Opt. Move). Qwen3-4B and Ministral3-3B show rough alignment between mention count and win\%, while interestingly mention count and execution \% aren't necessarily aligned.
}
\label{tab:nim_language_skills}
\end{table}

\paragraph{Nim}
Nim admits a complete algorithmic solution, related to a concept called "Nim-sum". We therefore test whether models possess knowledge of this strategy and whether they can execute it through the different language interfaces. Tab.~\ref{tab:nim_language_skills} presents per-language game log mentions of the optimal strategy. We use named mentions as a proxy for optimal strategy knowledge and find high variance across languages, with Qwen3-4B and Ministral3-3B showing rough alignment between optimal strategy mentions and win \%, where high-mention languages perform better than low-mention languages.

We note that knowledge of the optimal strategy does not necessarily translate into the ability to execute it, as executing the nim-sum strategy requires mathematical skill that may be disjoint from its knowledge. We therefore look for first-player first moves- which given our board definition, allow only one optimal move. We find that while Gemma-4-E4B-it is consistently able to execute the optimal move, Qwen3-4B and Ministral3-3B show a more complex pattern. We find dramatic differences between strategy mentions and successful execution; while English and French logs contain a similar number of strategy mentions for Qwen3-4B, English executes the optimal first move 80.8\% of the time while French does so only 24.6\% of the time; similar patterns can be found across languages for both models. This result hints at substantial and differing knowledge and mathematical reasoning gaps between languages.

Analyzing languages with a low number of strategy mentions reveals an interesting effect. For Ministral3-3B, 70\% of Arabic player optimal strategy mentions and 50\% of Hebrew player optimal strategy mentions originated from game logs where the model naturally language-switched into a Latin script. While this language switching isn't very common (3.7\% of Arabic logs and 1\% of Hebrew logs), it is responsible for many of the optimal strategy mentions of these language interfaces, implying that the differences in knowledge may even be larger than what we present. Moreover, these results showcase that even under the exact same game interface, simply switching the processing language can retrieve crucial knowledge that otherwise would have been lost, and directly tie into the recovery strategy from Sec.~\ref{sec:recovery}.

\end{document}